\documentclass{article}

\usepackage{arxiv}

\usepackage[utf8]{inputenc} % allow utf-8 input
\usepackage[T1]{fontenc}    % use 8-bit T1 fonts
\usepackage{hyperref}       % hyperlinks
\usepackage{url}            % simple URL typesetting
\usepackage{booktabs}       % professional-quality tables
\usepackage{amsfonts}       % blackboard math symbols
\usepackage{amsmath}
\usepackage{nicefrac}       % compact symbols for 1/2, etc.
\usepackage{microtype}      % microtypography
\usepackage{lipsum}		% Can be removed after putting your text content
\usepackage{graphicx}
\usepackage{natbib}
\usepackage{doi}
\usepackage{adjustbox}
\usepackage{tabularx}
\usepackage{booktabs}
\usepackage[dvipsnames]{xcolor}
\title{Bielik 11B v2 Technical Report}

\author{
 \textbf{Krzysztof Ociepa\textsuperscript{1,4}},
 \textbf{Łukasz Flis\textsuperscript{1,2}},
 \\
 \textbf{Remigiusz Kinas\textsuperscript{1}},
 \textbf{Krzysztof Wróbel\textsuperscript{1,3,5}},
 \textbf{Adrian Gwoździej\textsuperscript{1,2}}
\\
\\
 \textsuperscript{1}SpeakLeash,
 \textsuperscript{2}ACK Cyfronet AGH,
 \textsuperscript{3}Jagiellonian University,
 \textsuperscript{4}Azurro,
 \textsuperscript{5}Enelpol
}

\date{}

\renewcommand{\undertitle}{The Bielik LLM Team}
\renewcommand{\shorttitle}{Bielik 11B v2}

\hypersetup{
pdftitle={Bielik 11B v2},
pdfsubject={q-bio.NC, q-bio.QM},
pdfauthor={Krzysztof Ociepa, Łukasz Flis, Remigiusz Kinas, Krzysztof Wróbel, Adrian Gwoździej},
}

\begin{document}
\maketitle

\begin{abstract}
We present Bielik 11B v2, a state-of-the-art language model optimized for Polish text processing. Built on the Mistral 7B v0.2 architecture and scaled to 11B parameters using depth up-scaling, this model demonstrates exceptional performance across Polish language benchmarks while maintaining strong cross-lingual capabilities. We introduce two key technical innovations: Weighted Instruction Cross-Entropy Loss, which optimizes learning across diverse instruction types by assigning quality-based weights to training examples, and Adaptive Learning Rate, which dynamically adjusts based on context length. Comprehensive evaluation across multiple benchmarks demonstrates that Bielik 11B v2 outperforms many larger models, including those with 2-6 times more parameters, and significantly surpasses other specialized Polish language models on tasks ranging from linguistic understanding to complex reasoning. The model's parameter efficiency and extensive quantization options enable deployment across various hardware configurations, advancing Polish language AI capabilities and establishing new benchmarks for resource-efficient language modeling in less-represented languages.
\end{abstract}

% keywords can be removed
%%\keywords{First keyword \and Second keyword \and More}

\section{Introduction}

The rapid advancement in natural language processing (NLP) has led to the development of increasingly sophisticated language models that can understand and generate human-like text. These models have shown remarkable success in various linguistic tasks across multiple languages. However, the development of high-performing models for less-resourced languages remains a significant challenge due to the scarcity of large and diverse datasets and computational resources.

Several notable efforts have advanced Polish language modeling in recent years. TRURL 2, a collection of fine-tuned Llama 2 models with 7 billion and 13 billion parameters, was trained on approximately 1 million conversational samples. Qra models, comprising continuously pretrained architectures with 1, 7, and 13 billion parameters, leveraged 90 billion tokens of Polish data. More recently, PLLuM, developed by a consortium of Polish academic institutions, introduced models ranging from 8 billion to 70 billion parameters, created through continued pretraining of Llama and Mistral models on Polish corpora. While these initiatives have made important strides, many still face limitations in performance, versatility, or accessibility due to restrictive licensing or require significantly larger computational resources for comparable performance.

Building on our previous work with Bielik 7B v0.1, we introduce Bielik 11B v2, a state-of-the-art language model optimized specifically for Polish text processing. Developed as a collaborative effort between the SpeakLeash open-science project and ACK Cyfronet AGH, this model represents a significant advancement in both scale and capabilities. Bielik 11B v2 is built on the Mistral 7B v0.2 architecture and scaled to 11 billion parameters using depth up-scaling, striking an optimal balance between performance and computational efficiency.

Our approach introduces two key technical innovations: Weighted Instruction Cross-Entropy Loss, which optimizes learning across diverse instruction types by assigning quality-based weights to training examples, and Adaptive Learning Rate, which dynamically adjusts based on context length. These techniques, combined with comprehensive training on a diverse corpus of 198 billion tokens, enable Bielik 11B v2 to outperform many larger models, including those with 2-6 times more parameters, across a variety of benchmarks such as the Open PL LLM Leaderboard, Polish MT-Bench, and Polish Linguistic and Cultural Competency Benchmark (PLCC).

In the following sections, we detail the architecture of Bielik 11B v2, describe our dataset preparation methodology, discuss the pre-training and post-training processes, and evaluate the model's performance across multiple benchmarks. We also present various quantization options that enable deployment across different hardware configurations. Our results demonstrate that Bielik 11B v2 not only advances the state of Polish language understanding but also establishes new benchmarks for parameter-efficient language modeling in less-represented languages.

\section{Architecture}

\begin{table}[h]
\centering
\begin{tabular}{ll}
\toprule
Parameter & Value \\
\midrule
Layers & 50 \\
Model Dimension & 4096 \\
Attention Heads & 32 \\
Key/Value Heads & 8 \\
Head Size & 128 \\
Intermediate Size & 14336 \\
Activation Function & SwiGLU \\
Vocabulary Size & 32128 \\
Positional Embeddings & RoPE ($\theta = 1000000$) \\
Context Length & 32768 \\
\bottomrule
\end{tabular}
\caption{Model architecture.}
\label{tab:model-architecture}
\end{table}

\noindent The Bielik 11B v2 model is based on the Transformer architecture \cite{Vaswani2017AttentionIA}, with key parameters listed in Table \ref{tab:model-architecture}. The design integrates several advanced techniques to enhance performance and efficiency.

\noindent \textbf{Self-attention with causal masks} \cite{Vaswani2017AttentionIA} enables the model to assign varying importance to different parts of the input sequence. The causal mask ensures that the model only attends to preceding tokens, preserving the autoregressive property essential for language modeling.

\noindent \textbf{Grouped-query attention (GQA)} \cite{ainslie-etal-2023-gqa} reduces both computational complexity and memory usage while maintaining model quality. It achieves this by using fewer key-value heads than query heads, enabling more efficient handling of long sequences.

\noindent \textbf{SwiGLU activation function} \cite{Dauphin2016LanguageMW,shazeer2020gluvariantsimprovetransformer} combines the Swish activation function with Gated Linear Units (GLU), providing better performance and trainability than traditional activation functions such as ReLU.

\noindent \textbf{Rotary Positional Embeddings (RoPE)} \cite{SU2024127063} enhance the model's ability to capture relative token positions. Compared to absolute positional embeddings, RoPE supports better generalization to longer sequences and improves performance in tasks requiring positional sensitivity.

\noindent \textbf{Root Mean Square Layer Normalization (RMSNorm)} \cite{10.5555/3666122.3668105} normalizes activations within the network, offering greater training stability and slightly faster computation compared to standard Layer Normalization.

\noindent \textbf{Pre-normalization} involves applying layer normalization before the self-attention and feed-forward layers. This improves model convergence and overall performance.

\begin{figure}[ht]
\centering
\includegraphics[width=\columnwidth]{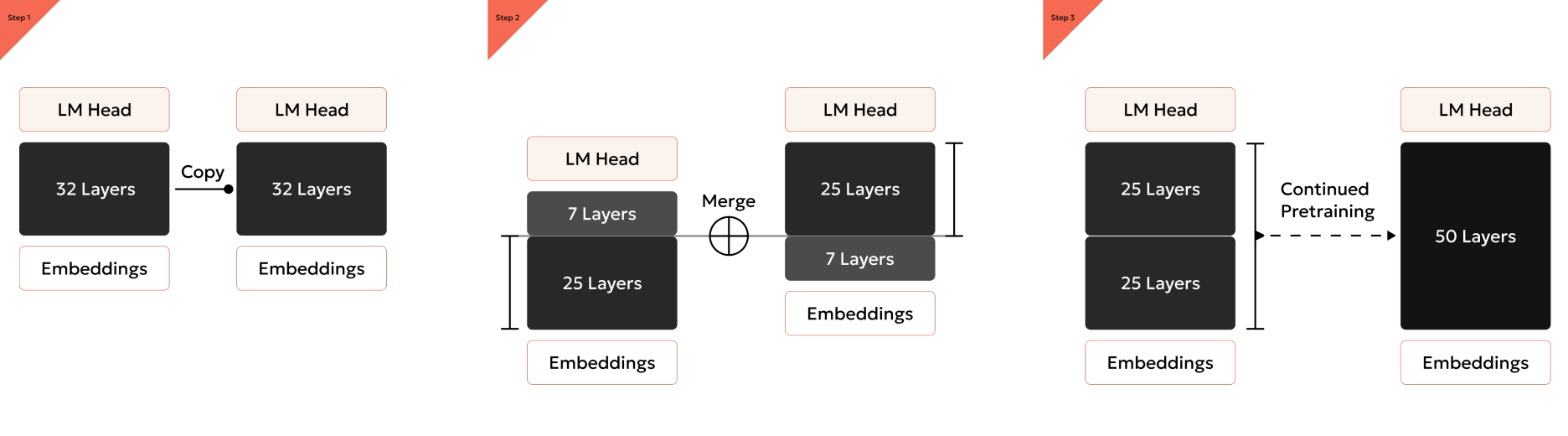}
\caption{Depth up-scaling with $n = 32$, $m = 7$, and $s = 50$.}
\label{fig:model-upscaling}
\end{figure}

The Bielik 11B v2 model is adapted from the Mistral 7B v0.2 model \cite{jiang2023mistral7b}, scaled using the Depth Up-Scaling method \cite{kim2024solar107bscalinglarge}, and further pretrained. Beginning with the original 32-layer model ($n = 32$), we duplicated its layers and removed the final 8 and initial 8 layers at the junction ($m = 7$), resulting in a 50-layer model ($s = 50$), as shown in Figure \ref{fig:model-upscaling}. We chose to upscale to 50 layers in order to reach a model size of approximately 11B parameters, which can still be run comfortably on consumer-grade GPUs with up to 24GB of VRAM.

This decision to build on an existing model, rather than developing one from scratch, was driven by the desire to allocate resources efficiently. By focusing on the linguistic adaptation of an already high-performing model, we were able to optimize both time and computational resources. The Mistral 7B v0.2 model was selected due to its strong benchmark performance and permissive Apache 2.0 license.

We retained the original tokenizer from the Mistral 7B model, which includes a vocabulary of 32,000 tokens. The only modification was the addition of special tokens required for the chat template, bringing the total vocabulary size to 32,128 tokens.

\section{Pre-training}

The primary objective of the pre-training phase was to enhance the model's proficiency in the Polish language, with an emphasis on both accuracy and fluency. To achieve this, we utilized a diverse collection of high-quality Polish texts. These materials underwent rigorous preprocessing and thorough quality evaluation to ensure the highest standards of training data.

\subsection{Pre-training Data} \label{Pre-training-Data}
The pre-training of the Bielik v2 model involved constructing a novel, diverse, and high-quality dataset composed primarily of Polish-language texts. We leveraged resources from the SpeakLeash project \cite{speakleashorg}. Using metadata associated with each document—including topical information and various stylometric features—we selected 41 million documents from different datasets, ensuring both high quality and thematic diversity. These selected texts underwent comprehensive cleaning and quality evaluation, as described in Sections \ref{Data-Cleanup} and \ref{Quality-Evaluation}. 

Additionally, we excluded documents where scraping was technically permitted (i.e., not blocked by robots.txt) but where the terms and conditions explicitly prohibited use for training language models. Only documents meeting our stringent quality standards were retained and subsequently tokenized. This meticulous curation resulted in a Polish training corpus of 90 billion tokens.

To improve the model's adaptation to Polish while mitigating catastrophic forgetting \cite{Li2022OvercomingCF,pmlr-v199-ostapenko22a,ibrahim2024simplescalablestrategiescontinually}, we supplemented the dataset with English texts from the SlimPajama dataset \cite{cerebras2023slimpajama}, known for its diversity and quality.

To better align the model with the objectives of later training phases, we also incorporated the instruction dataset described in Section~\ref{Post-training-Data} into the pre-training data. Although originally constructed for supervised fine-tuning (SFT), this dataset facilitated a smoother and more effective transition to subsequent stages of training.

In total, the final training dataset comprised 198 billion tokens (96 million documents).

\subsubsection{Data Cleanup} \label{Data-Cleanup}
To enhance the quality of the documents, we applied a series of heuristics designed to remove corrupted or irrelevant content, anonymize personal data (including physical addresses, email addresses, phone numbers, and URLs), and resolve encoding or formatting issues. These steps produced cleaner, higher-quality texts that were subsequently subjected to further evaluation.

\subsubsection{Quality Evaluation}
\label{Quality-Evaluation}

To develop the training dataset for text quality evaluation, we manually selected and annotated documents, categorizing them into three quality classes: \textbf{HIGH}, \textbf{MEDIUM}, and \textbf{LOW}. The \textbf{HIGH} class represents superior-quality documents; \textbf{LOW} denotes poor-quality texts, and \textbf{MEDIUM} encompasses documents whose quality is ambiguous, falling between high and low standards. This nuanced classification approach addresses the inherent complexities of assessing textual quality.

For the Bielik v2 model, the annotated dataset was significantly expanded, comprising \textbf{20\,000} training samples, \textbf{1\,500} test samples, and \textbf{500} validation samples. Each document was represented by a vector of \textbf{150} carefully selected stylometric features. Beyond standard linguistic and structural metrics (such as frequencies of verbs, nouns, sentences, and punctuation marks), the updated feature set placed special emphasis on characteristics relevant to Markdown-formatted texts.

Newly introduced Markdown-focused features include: counts of ordered list items, hyperlinks, images, inline code fragments, code blocks, blockquotes, horizontal rules, and special Markdown characters. In addition, detailed analysis of table formatting—such as pipe character frequency and header-to-body alignment—was incorporated.

This comprehensive feature set was developed following the methodology inspired by the StyloMetrix tool~\cite{okulska2023stylometrixopensourcemultilingualtool}, and enhanced specifically for Markdown textual analysis. Using this enriched stylometric and Markdown-aware representation, we evaluated multiple machine learning algorithms, with the \textbf{XGBoost} classifier emerging as the most effective for distinguishing between the defined quality categories.

A ranked list of the most influential features, based on mean absolute SHAP values, is presented in Table~\ref{tab:shap}.

The performance of the Bielik v2 model was rigorously evaluated on held-out validation and test sets. On the validation set, the model achieved an overall \textbf{accuracy of 86\%} and a \textbf{macro-average F1-score of 0.79}, indicating robust performance across all three quality levels. On the test dataset, the classifier reached an overall \textbf{accuracy of 94\%} and a \textbf{macro-average F1-score of 0.85}, underscoring its strong ability to differentiate particularly between HIGH- and LOW-quality documents.

To determine a reliable threshold for identifying high-quality texts suitable for inclusion in downstream training corpora, we conducted a manual analysis of \textbf{1\,000} documents. This evaluation confirmed that a predicted probability above \textbf{90\%} for the HIGH-quality class serves as a practical cutoff. Documents below this threshold were systematically excluded from the final training dataset for the Bielik v2 model.

\paragraph{Alternative high-capacity model.}
The \texttt{XGB\_HighN\_LowLR\_d7\_n1000\_lr002} configuration corresponds to an \texttt{XGBoostClassifier} trained with 
\texttt{n\_estimators = 1000}, \texttt{max\_depth = 7}, \texttt{learning\_rate = 0.02}, \texttt{subsample = 0.7}, and 
\texttt{colsample\_bytree = 0.7}. As with the best model, it uses the \texttt{multi:softprob} objective, 
with \texttt{eval\_metric = mlogloss}, \texttt{use\_label\_encoder = False}, and a fixed \texttt{random\_state = 42}.

\begin{table}[htbp]
\centering
\begin{adjustbox}{max width=\textwidth}
\begin{tabularx}{\textwidth}{lccc}
\toprule
\textbf{Model} & \textbf{Val F1 (macro)} & \textbf{Val F1 (weighted)} & \textbf{Val Accuracy} \\
\midrule
XGB\_HighN\_LowLR\_d7\_n1000\_lr002 & 0.789482 & 0.928575 & 0.938 \\
XGB\_AggressiveSubsample\_d8\_n300\_lr007 & 0.796013 & 0.928040 & 0.934 \\
LightGBM & 0.798175 & 0.927704 & 0.934 \\
XGB\_RegL1L2\_d8\_n250\_lr008\_a02\_l05 & 0.793790 & 0.926026 & 0.934 \\
CatBoost & 0.788047 & 0.924801 & 0.932 \\
HistGradientBoosting & 0.788035 & 0.924784 & 0.932 \\
XGBoost\_nEstimators400\_maxDepth3\_lr025 & 0.791143 & 0.924473 & 0.930 \\
XGBoost & 0.763592 & 0.917313 & 0.926 \\
MLP\_hidden100\_relu\_adam & 0.768963 & 0.913112 & 0.918 \\
MLP\_hidden50\_50\_tanh\_sgd & 0.741196 & 0.905938 & 0.914 \\
EBM & 0.725163 & 0.904523 & 0.916 \\
ExtraTrees\_nEstimators200\_maxDepthNone & 0.730563 & 0.901927 & 0.920 \\
RandomForest\_nEstimators200\_maxDepthNone & 0.729183 & 0.899997 & 0.918 \\
MLP\_hidden200\_tanh\_adam\_alpha1e-3 & 0.746473 & 0.898842 & 0.900 \\
RandomForest\_nEstimators300\_maxDepth20 & 0.727778 & 0.898033 & 0.916 \\
RandomForest\_nEstimators150\_maxDepth15 & 0.724613 & 0.896521 & 0.914 \\
MLP\_hidden50\_25\_10\_logistic\_adam & 0.741392 & 0.895995 & 0.896 \\
MLP\_hidden100\_50\_relu\_lbfgs & 0.739666 & 0.895068 & 0.898 \\
RandomForest\_nEstimators100\_maxDepth10 & 0.669666 & 0.896601 & 0.908 \\
KNN\_neighbors7\_weightsUniform\_norm1 & 0.750744 & 0.884859 & 0.902 \\
KNN\_neighbors5\_weightsDistance\_norm1 & 0.732194 & 0.880066 & 0.884 \\
\bottomrule
\end{tabularx}
\end{adjustbox}
\vspace{6pt}
\caption{Comparison of models performance}
\end{table}

Using these features as input, we trained an XGBoost classifier. This machine learning model effectively captured complex patterns in the data, achieving strong results on both validation and test sets.

Detailed performance metrics for the validation set are presented in Table~\ref{tab:classification-validation}, while results for the test set are summarized in Table~\ref{tab:classification-test}.

\begin{figure}[ht]
\centering
\includegraphics[width=\columnwidth]{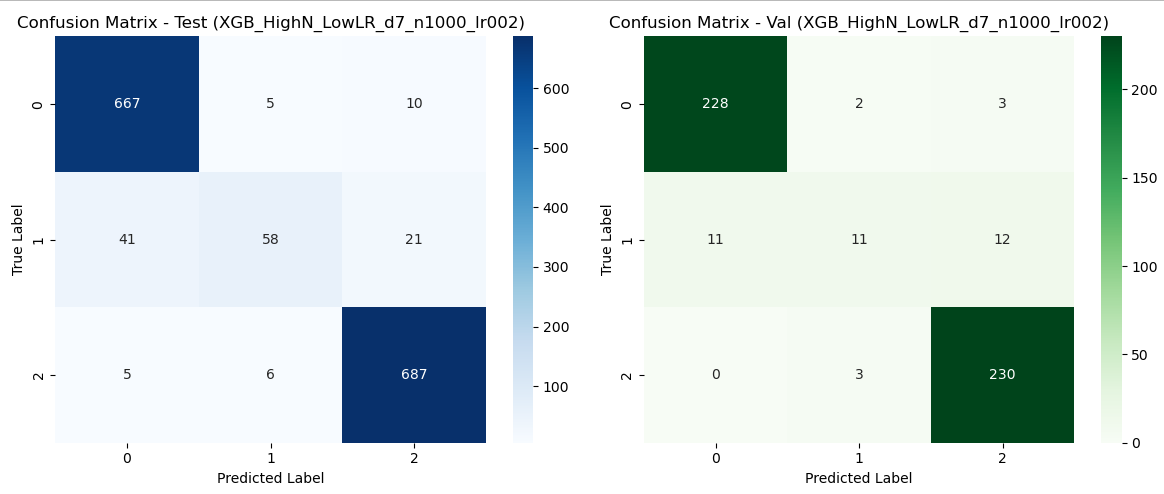}
\caption{Confusion matrix showing test and validation results for the XGBoost classifier.}
\label{fig:XGBoost-validation}
\end{figure}

\begin{table}[htbp]
\centering
\begin{tabular}{lcccc}
\toprule
Class & Precision & Recall & F1-score & Support \\
\midrule
LOW & 0.95 & 0.98 & 0.97 & 233 \\
MEDIUM & 0.69 & 0.32 & 0.44 & 34 \\
HIGH & 0.94 & 0.99 & 0.96 & 233 \\
\midrule
\textbf{Accuracy} & \multicolumn{3}{c}{0.94} & 500 \\
\textbf{Macro avg} & 0.86 & 0.76 & 0.79 & 500 \\
\textbf{Weighted avg} & 0.93 & 0.94 & 0.93 & 500 \\
\bottomrule
\end{tabular}
\vspace{6pt}
\caption{Classification Report (Validation) – Best Model}
\label{tab:classification-validation}
\end{table}

\begin{table}[htbp]
\centering
\begin{tabular}{lcccc}
\toprule
Class & Precision & Recall & F1-score & Support \\
\midrule
LOW & 0.94 & 0.98 & 0.96 & 682 \\
MEDIUM & 0.84 & 0.48 & 0.61 & 120 \\
HIGH & 0.96 & 0.98 & 0.97 & 698 \\
\midrule
\textbf{Accuracy} & \multicolumn{3}{c}{0.94} & 1500 \\
\textbf{Macro avg} & 0.91 & 0.82 & 0.85 & 1500 \\
\textbf{Weighted avg} & 0.94 & 0.94 & 0.94 & 1500 \\
\bottomrule
\end{tabular}
\vspace{6pt}
\caption{Classification Report (Test) – Best Model}
\label{tab:classification-test}
\end{table}

To determine an appropriate threshold for selecting high-quality documents, we conducted a manual analysis of 1,000 samples. Based on this evaluation, we set the threshold for the HIGH category at a predicted probability exceeding 90\%. Documents falling below this threshold were excluded from the Bielik model's final training dataset.

\begin{table}[htbp]
\centering
\begin{tabular}{lc}
\toprule
\textbf{Feature} & \textbf{Mean Abs SHAP} \\
\midrule
oovs & 0.5637 \\
non\_alpha\_word\_fractions & 0.3562 \\
stop\_word\_ratio & 0.3155 \\
average\_lines & 0.2147 \\
colons\_per\_sentence & 0.1583 \\
short\_line\_ratio\_20 & 0.1578 \\
special\_chars\_ratio\_md & 0.1570 \\
commas\_per\_sentenced & 0.1021 \\
lowercase\_ratio\_md & 0.1003 \\
overall\_uppercase\_ratio & 0.1002 \\
avg\_paragraph\_length & 0.0984 \\
avg\_word\_length & 0.0955 \\
noun\_ratio & 0.0906 \\
char\_ratio\_\_ & 0.0906 \\
adj\_ratio & 0.0904 \\
uppercase\_ratio\_md & 0.0838 \\
ratio\_of\_bulletpoints & 0.0822 \\
char\_ratio\_> & 0.0816 \\
punct\_frequency & 0.0804 \\
duplicate\_line\_ratio & 0.0788 \\
blank\_lines & 0.0771 \\
emoticons & 0.0754 \\
single\_char\_ratio & 0.0711 \\
entropy & 0.0710 \\
blank\_lines\_ratio & 0.0705 \\
\bottomrule
\end{tabular}
\vspace{6pt}
\caption{Top Features by Mean Absolute SHAP Value}
\label{tab:shap}
\end{table}

\subsubsection{Category Classification: Results and Applications}
\label{Category-Classification}

This section discusses the performance and practical implications of the text category classifier, designed to automatically label Polish-language documents with one of 120 predefined thematic categories. The classifier plays a crucial role not only in organizing data for training base models but also in generating balanced instruction sets covering diverse domains.

\subsubsection*{Dataset and Setup}
The classification model was trained on a comprehensive dataset of 35,944 Polish-language documents. A standard 90/10 train-test split was applied, resulting in 32,349 samples for training and 3,595 for evaluation.

To ensure the representativeness of the full label set, stratified sampling was employed, preserving the category distribution across the two subsets. This approach is especially important for tasks involving a large number of categories, some of which may be infrequent or closely related.

\subsubsection*{Modeling Approach}
The classification pipeline starts by converting textual data into numeric vectors using a \texttt{CountVectorizer}, restricted to the 10,000 most informative and frequently occurring features. This dimensionality has proven sufficient for effectively separating the 120 categories.

The term counts were then transformed into TF-IDF representations via the \texttt{TfidfTransformer}, highlighting the relative importance of terms across documents.

Classification was performed using a \texttt{Linear Support Vector Classifier} (LinearSVC), known for its efficiency and robustness on high-dimensional data. To produce calibrated probability estimates, the classifier was wrapped with \texttt{CalibratedClassifierCV} using isotonic regression and 3-fold cross-validation.

\subsubsection*{Performance Evaluation}
The trained model achieved an overall accuracy of \textbf{94.44\%} on the held-out test set of 3,595 documents, demonstrating strong generalization across the wide thematic range.

These consistently high scores indicate that the classifier performs reliably across both high- and low-frequency classes. Importantly, most misclassifications tend to occur among semantically related categories (e.g., \textit{Transportation}, \textit{Automotive}, and \textit{Travel}), which are not problematic in practical downstream use.

\subsubsection*{Category Distribution}

\begin{figure}[!htbp]
\centering
\includegraphics[width=\columnwidth]{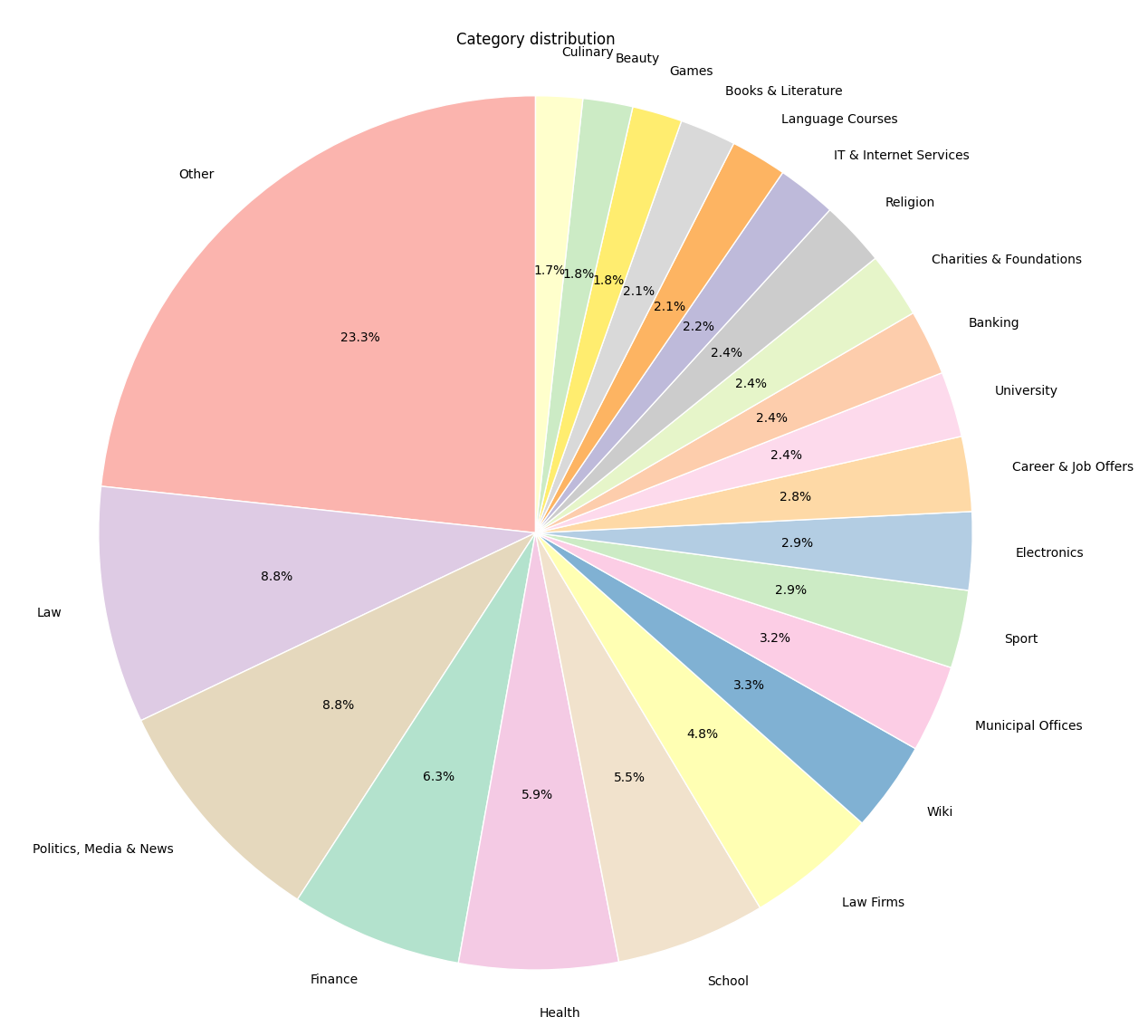}
\caption{Distribution of major thematic categories in the Polish text dataset ($\geq1.7\%$)}
\label{fig:distribution}
\end{figure}

The underlying dataset covers a broad range of topics, with category distribution reflecting real-world domain diversity. As shown in Figure~\ref{fig:distribution}, the most prominent categories include \textit{Other} (23.3\%), \textit{Law} (8.8\%), and \textit{Politics, Media \& News} (8.8\%), followed by domains such as \textit{Finance}, \textit{Health}, and \textit{Education}. 

While the long-tail nature of the distribution is evident, the dataset remains well-balanced overall. Categories with lower representation (e.g., below 1.5\%) were excluded from the analysis in the figure to improve readability, though they were retained in the training process. This distribution ensures the classifier is exposed to sufficient variability, while preserving robustness even across less frequent or closely related classes.

\begin{table}[htbp]
\centering
\label{tab:category-classification-results}
\begin{tabular}{lcccc}
\toprule
Metric & Precision & Recall & F1-score & Support \\
\midrule
\textbf{Macro avg} & 0.9451 & 0.9444 & 0.9436 & 3595 \\
\textbf{Weighted avg} & 0.9447 & 0.9441 & 0.9426 & 3595 \\
\bottomrule
\end{tabular}
\vspace{6pt}
\caption{Category Classification Performance on Test Set (N=3,595)}
\end{table}

\subsubsection*{Application and Future Directions}
This classifier enables the construction of thematically diverse and well-balanced datasets in the Polish language, which are essential for both model pretraining and the generation of high-quality, targeted instruction sets.

Its ability to maintain category balance while minimizing critical misclassifications allows it to serve as a robust preprocessing step for synthetic instruction generation. These instructions benefit from broad topical coverage and linguistic quality, strengthening downstream models' performance in instruction-following tasks across various domains.

\subsection{Training Hyperparameters}
We employed the AdamW optimizer \cite{Loshchilov2017DecoupledWD} with hyperparameters $\beta_1 = 0.9$, $\beta_2 = 0.95$, and a weight decay of 0.1. The learning rate followed a cosine decay schedule, starting at $2 \times 10^{-5}$ and decreasing to $9 \times 10^{-6}$, with a warm-up period of 50 iterations. Training was conducted over 97,250 iterations.

We utilized Megatron-LM with a global batch size of 128 and a tensor parallelism degree of 4. The gradient clipping norm was set to 1.0, and mixed-precision training was enabled using bfloat16. The model was trained on 198 billion tokens over two epochs, with a maximum context length of 32,768 tokens.

\subsection{Training process monitoring}
To ensure the expected quality of the pre-training process, we carefully monitored the progression of model performance on the OpenLLM PL and OpenLLM EN benchmarks. Throughout training, model checkpoints were saved at regular intervals and benchmarked accordingly. In our experience, degradation in benchmark performance over time typically indicates dataset-related issues and is rarely reversible. 

Performance evaluation based on the benchmarks mentioned above was conducted in parallel with the training process on a separate set of compute nodes to avoid interference. This approach enabled early detection of training issues without having to wait for an epoch to complete, resulting in significant compute time savings.

\section{Post-training}
After completing the pre-training phase, we proceeded to the post-training phase, which aimed to enhance the model's performance in various domains, including coding, mathematics, logical reasoning, and instruction following.

\subsection{Post-training Data} \label{Post-training-Data}
Due to the lack of a sufficiently large and open dataset of instructions and dialogues in the Polish language, we began constructing our own dataset, which is continuously expanded and refined by human annotators. This dataset is manually curated through the creation of instructions and dialogues, ensuring high-quality, relevant content specifically tailored to the Polish language.

To complement the manually annotated data, we also generated additional instructions and dialogues using Mixtral 8x22B \cite{jiang2024mixtralexperts}. The resulting dataset used for training included over 20 million instructions, totaling more than 10 billion tokens. 

\begin{figure}[!htbp]
\centering
\includegraphics[width=\columnwidth]{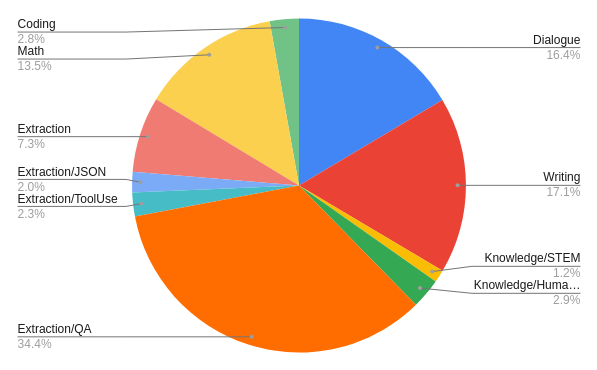}
\caption{Distribution of instruction categories in the SFT dataset.}
\label{fig:sft-category-distribution}
\end{figure}

\begin{figure}[!htbp]
\centering
\includegraphics[width=\columnwidth]{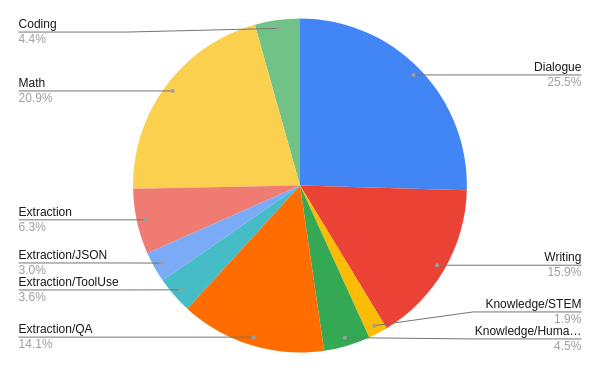}
\caption{Weighted distribution of instruction categories in the SFT dataset.}
\label{fig:sft-category-distribution-weighted}
\end{figure}

The distribution of instructions and dialogues across categories is presented in Figure~\ref{fig:sft-category-distribution}, and the weighted version, based on the weights described in Section~\ref{WICEL}, is shown in Figure~\ref{fig:sft-category-distribution-weighted}.

For the final phase of post-training with reinforcement learning using Direct Preference Optimization (DPO) \cite{rafailov2024directpreferenceoptimizationlanguage} and (DPO-P) \cite{pal2024smaugfixingfailuremodes}, we constructed a dedicated dataset consisting of 72,000 preference instructions. To support alignment training, we developed a comprehensive instruction data generation pipeline designed to ensure high-quality, diverse, and reliable datasets in Polish. The pipeline combines manual efforts with automated processes, incorporating techniques to maximize task variety and rigorously control data quality throughout each stage.
\begin{itemize}
\item \textbf{Step 1:} Instruction Creation.
The instruction creation phase involved two approaches. First, domain experts manually authored instructions across a wide range of topics and complexity levels. Second, a perturbation-based augmentation strategy was applied: existing English-language instruction datasets were translated and perturbed into Polish using language models. This allowed us to expand the instruction pool by introducing meaningful linguistic and semantic variations while preserving the core task intent.

\item \textbf{Step 2:} Instruction Deduplication.
To increase diversity and minimize redundancy, all instructions were subjected to a deduplication process. We employed a combination of Locality-Sensitive Hashing (LSH), cosine similarity calculations, and MiniHash fingerprinting techniques. Depending on the configuration selected by domain experts, these methods were applied individually or in combination, ensuring that similar or duplicate instructions were effectively removed from the dataset.

\item \textbf{Step 3:} Generation of Chosen and Rejected Response Pairs.
Following instruction preparation, we generated pairs of preferred (chosen) and less preferred (rejected) responses. This was achieved through two main methods: manual labeling by human annotators and synthetic generation using large language models, including Mistral, Bielik v0.1, and LLaMA-based models. Manual creation allowed for fine-grained quality distinctions, while automated generation enabled efficient scaling of the dataset.

\item \textbf{Step 4:} Quality Evaluation with Metamodels.
Each response pair was evaluated using reward metamodels based on the LLaMA 70B and Mixtral8x22B. The metamodels assigned quality scores on a scale from 0 to 10, considering attributes such as factual accuracy, completeness, clarity, and depth of reasoning. This evaluation provided an essential quantitative assessment of response quality for subsequent filtering.

\item \textbf{Step 5:} Filtering.
The dataset was filtered according to strict quality control rules. Pairs in which the chosen and rejected responses received identical scores were discarded. Likewise, pairs with a score distance of less than two points were removed to avoid ambiguous training examples. Additionally, if a rejected response achieved a higher score than the chosen response, the labels were swapped to maintain logical consistency.

\item \textbf{Step 6:} Dataset Cleaning.
After filtering, the dataset underwent a cleaning phase to ensure format consistency and correctness. Samples were reformatted into a unified structure, and entries containing API errors, malformed outputs, or serious formatting issues were removed. This stage ensured the structural and semantic integrity of the dataset.

\item \textbf{Step 7:} Manual Inspection and Correction.
Finally, the cleaned dataset was subjected to a full manual review. Annotators validated the quality of instruction-response pairs, corrected mislabeled examples, and adjusted evaluation scores where necessary. This manual inspection phase was crucial for maximizing dataset reliability and ensuring its suitability for downstream RLHF training.
\end{itemize}

It is important to note that our alignment objective was not centered on safety-related aspects of model responses. Instead, our focus was on tailoring the model to produce outputs in a preferred style, encompassing appropriate dialogue formatting, mathematical expressions, storytelling, bullet points, headings, and other stylistic elements. To achieve this, we endeavored to create a highly diversified set of tasks and, subsequently, to align the selected responses to a consistent stylistic standard.

Furthermore, the range of tasks included in the instruction dataset is continuously expanding. New tasks are generated based on the ongoing observation of user interactions with the Bielik Chat system. By analyzing real-world conversations, we identify gaps, emerging user needs, and opportunities for stylistic or functional improvements, which in turn inform the creation of additional instructions and response patterns. This iterative approach ensures that the dataset evolves dynamically, staying aligned with user expectations and practical usage scenarios.

\subsection{Supervised Fine-Tuning}
Previous research has shown that low-quality instructions negatively impact a model's benchmark performance, with findings indicating that poorly constructed instructions degrade model capabilities \cite{NEURIPS2023_ac662d74}. These studies demonstrated that smaller, high-quality instruction datasets often outperform larger, noisier ones. To mitigate this issue, we implemented several improvements, as summarized below, while continuing to use the aforementioned datasets.

\subsubsection{Masked Tokens}
We adopted a masked token strategy, applying the loss function selectively to specific parts of the output. In particular, we masked the loss for user instructions and control tokens \cite{shi2024instructiontuninglossinstructions}. This method ensures that these tokens do not contribute to the total loss during training, allowing the model to concentrate on learning from the actual content tokens.

\subsubsection{Adaptive Learning Rate} \label{ALR}
Instruction lengths can vary significantly, causing fluctuations in the number of tokens used to compute the loss. To ensure that each instruction had a consistent influence during training, we utilized an adaptive learning rate (ALR) \cite{ociepa2024bielik7bv01polish}, where the learning rate (LR) is scaled based on the square root of the ratio between the number of tokens in the current batch (T) and a baseline batch size (BS):
\begin{equation}
\text{ALR} = \text{LR} \cdot \sqrt{\frac{\text{T}}{\text{BS}}}
\end{equation}

\subsubsection{Weighted Instruction Cross-Entropy Loss} \label{WICEL}
To fine-tune our language model on a dataset containing instruction-response pairs of varying quality, we applied the Weighted Instruction Cross-Entropy Loss (WICEL) \cite{ociepa2024bielik7bv01polish}. This technique assigns a quality-based weight $w_i \in (0, 1]$ to each training example, enabling the model to prioritize higher-quality data during optimization. The weighted loss is calculated as:

\begin{equation}
\mathcal{L} = -w_i \cdot \sum_{t=1}^{T} \log \pi_\theta(y_t | x, y_{<t})
\end{equation}

where $(x, y)$ denotes the instruction-response pair, $T$ is the length of the response, and $\pi_\theta$ is the fine-tuned model. This approach increases training robustness and efficiency by incorporating lower-quality samples without allowing them to dominate the learning process.

\subsection{Supervised Fine-Tuning Hyperparameters}

We used the AdamW optimizer with hyperparameters $\beta_1 = 0.9$, $\beta_2 = 0.95$, and a weight decay of 0.05. The learning rate followed a cosine decay schedule, starting at $7 \times 10^{-6}$ and decaying to $6 \times 10^{-7}$, with 50 warmup iterations.

Our configuration used a global batch size of 128, composed of local batches of size 1. Gradient clipping was applied with a threshold of 1.0, and training was performed using mixed precision with the bfloat16 format.

We employed a sample packing technique, which concatenates multiple samples from the dataset into a single sequence, up to the maximum sequence length. The model was trained for 3 epochs with a maximum context length of 8,192 tokens.

\subsection{Reinforcement Learning}

To align the model with user preferences, we experimented with several techniques: DPO \cite{rafailov2023direct}, PPO\cite{schulman2017ppo}, KTO \cite{ethayarajh2024kto}, ORPO \cite{hong-etal-2024-orpo} and SiMPO \cite{meng2024simpo}. As part of the project, we also developed a Polish reward model based on preference data. However, training with Proximal Policy Optimization (PPO) using this reward model did not yield performance improvements over the DPO approach. Ultimately, the DPO-Positive (DPO-P) method \cite{pal2024smaugfixingfailuremodes} was adopted. 

Direct Preference Optimization (DPO) is a technique that aligns language models with human preferences by directly optimizing a loss function based on pairwise comparisons between preferred (chosen) and less preferred responses (rejected). However, a notable limitation of DPO is its potential to inadvertently decrease the absolute probability of preferred responses, especially when the differences between response pairs are minimal. This phenomenon arises because DPO focuses solely on the relative preference between responses, without explicitly encouraging the model to maintain or increase the likelihood of preferred outputs.

\begin{equation}
\mathcal{L}_{\text{DPO}}(\pi_\theta; \pi_{\text{ref}}) = - \mathbb{E}_{(x, y_{\text{w}}, y_{\text{l}}) \sim \mathcal{D}} \left[
\log \sigma \left(
\beta \left(
\log \frac{\pi_\theta(y_{\text{w}} \mid x)}{\pi_{\text{ref}}(y_{\text{w}} \mid x)} - \log \frac{\pi_\theta(y_{\text{l}} \mid x)}{\pi_{\text{ref}}(y_{\text{l}} \mid x)}
\right)
\right)
\right]
\end{equation}

To address this issue, DPO-Positive (DPO-P) introduces an additional term to the loss function that penalizes reductions in the probability of preferred responses relative to a reference model. This modification ensures that the model not only favors preferred responses over less preferred ones but also maintains or enhances the generation probability of these preferred responses. Empirical studies have demonstrated that DPO-P outperforms standard DPO, particularly in scenarios where response pairs have small edit distances, by mitigating the degradation of preferred response probabilities and improving overall model alignment with human preferences.

\begin{equation}
\mathcal{L}_{\text{DPOP}}(\pi_\theta; \pi_{\text{ref}}) = - \mathbb{E}_{(x, y_{\text{w}}, y_{\text{l}}) \sim \mathcal{D}} \left[
\log \sigma \left(
\beta \left(
\log \frac{\pi_\theta(y_{\text{w}} \mid x)}{\pi_{\text{ref}}(y_{\text{w}} \mid x)} - \log \frac{\pi_\theta(y_{\text{l}} \mid x)}{\pi_{\text{ref}}(y_{\text{l}} \mid x)}
\right)
- \lambda \cdot \max\left(0, \log \frac{\pi_{\text{ref}}(y_{\text{w}} \mid x)}{\pi_\theta(y_{\text{w}} \mid x)} \right)
\right)
\right]
\end{equation}

This approach leveraged both generated and manually corrected examples, which were scored by a metamodel. A dataset of over 72,000 examples, varying in length to capture different aspects of response style, was used. This dataset was filtered and assessed by the reward model to select instructions with an appropriate level of distinction between preferred and rejected responses. A key innovation introduced in DPO-P was the incorporation of multi-turn conversations.

\subsubsection{DPO-Positive Hyperparameters}

During DPO-Positive (DPO-P) training, the loss function was parameterized with $\beta = 0.05$ and $\lambda = 2.5$, following the recommendations for stabilizing preference optimization and ensuring retention of high-quality preferred responses.

We used the AdamW optimizer with hyperparameters $\beta_1 = 0.9$, $\beta_2 = 0.95$, and no weight decay. The learning rate was constant and set to $7 \times 10^{-7}$, with 50 warmup iterations. Training was conducted over a total of 3,800 iterations.

Our configuration used a global batch size of 64, composed of local batches of size 1. Gradient clipping was applied with a threshold of 1.0, and training was performed using mixed precision with the bfloat16 format.

\subsection{Model Merging}

To leverage the diverse capabilities of models fine-tuned under varying conditions, we developed our framework for merging models and tested multiple strategies, including Linear (Model Soups) \cite{wortsman2022modelsoups}, TIES (TIES-Merging: Resolving Interference When Merging Models) \cite{yadav2023tiesmerging}, Model Stock \cite{jang2024modelstock}. Each of these methods offers distinct approaches to combining model parameters, aiming to integrate strengths from different models while mitigating potential conflicts or redundancies. 

We systematically evaluated these merging techniques on a suite of Polish language benchmarks. Among the methods tested, the Linear merging approach, assigning equal weights (1.0) to each model, consistently yielded the most favorable results. This method involved merging three models obtained from different phases of the same DPO-Positive (DPO-P) training process, effectively capturing varied response characteristics developed during training.

To further explore the potential of model merging, we conducted an experiment by introducing the original base model into the mixture of the three DPO-P fine-tuned models. This addition aimed to assess whether incorporating the foundational model could enhance the merged model's performance or stability.

Beyond post-alignment merging, we also applied merging techniques during earlier stages of model development, specifically between different Supervised Fine-Tuning (SFT) runs. This strategy aimed to consolidate improvements from various SFT iterations, thereby enhancing the base model's quality prior to alignment through reinforcement learning from human feedback (RLHF).

\subsection{Layer-Wise Model Quality Monitoring}
To ensure the robustness and generalization capabilities of our models throughout the training process, we employed WeightWatcher \cite{martin2021implicit}, an open-source diagnostic tool designed to analyze deep neural networks without relying on training or test data. WeightWatcher examines the spectral properties of weight matrices in each layer, leveraging the theory of Heavy-Tailed Self-Regularization (HTSR) to assess layer quality.

\begin{figure}[!htbp]
\centering
\includegraphics[width=\columnwidth]{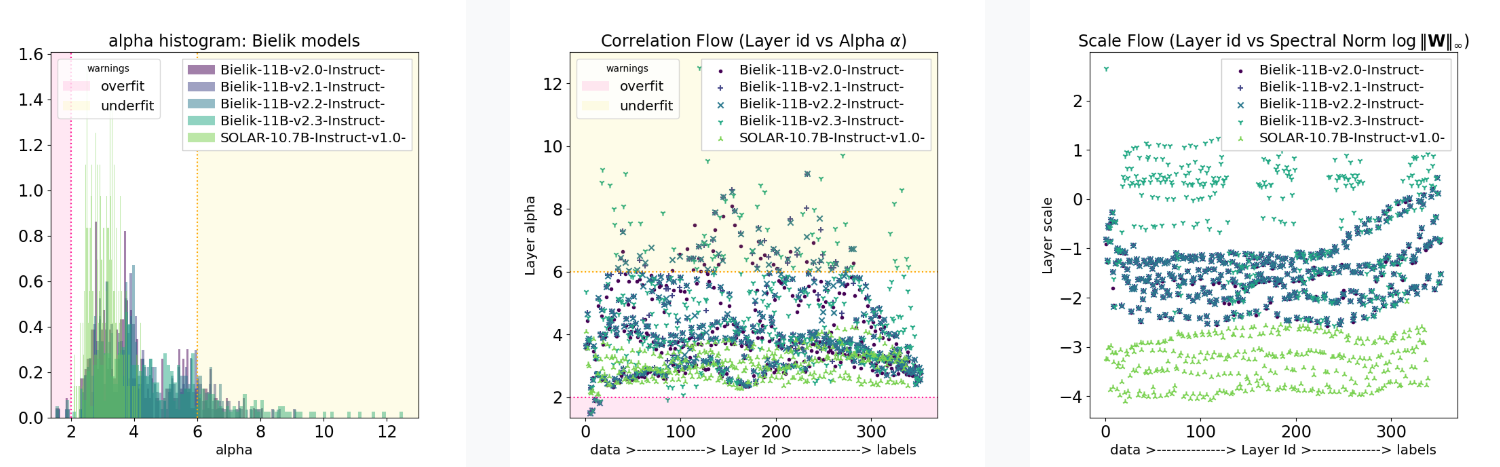}
\caption{Alpha Histogram, Correlation Flow and Scale Flow charts for Bielik-11B models.}
\label{fig:layer-wise-quality-monitoring}
\end{figure}

A key metric provided by WeightWatcher is the power-law exponent $\alpha$, which quantifies the degree of heavy-tailedness in the eigenvalue distribution of a layer's weight matrix. Empirical studies suggest that well-trained layers typically exhibit $\alpha$ values within the range of 2 to 6, with lower values indicating better generalization properties.

During our experiments, we applied WeightWatcher to each model checkpoint obtained from different phases of the DPO-Positive (DPO-P) training process. This analysis enabled us to identify layers that were potentially over- or under-trained, providing insights into the internal dynamics of the model's learning process. By monitoring the evolution of $\alpha$ values across layers and training stages, we could make informed decisions regarding training adjustments and model selection, ultimately contributing to the development of models with improved performance and stability.

\section{Quantization}

In order to support a wider range of hardware configurations and deployment scenarios, we provide several quantized variants of our instruction-tuned models. Quantization significantly reduces model size and inference latency, making it possible to run models on resource-constrained devices, albeit with a potential trade-off in output quality.

We have prepared the following quantized versions of our models:

\begin{itemize}
    \item GGUF
    \item GPTQ
    \item HQQ
    \item AWQ
    \item EXL2
    \item MLX
    \item Quanto
    \item FP8 (compatible with vLLM and SGLang; optimized for Ada Lovelace and Hopper architectures)
    \item INT8 W8A8 (vLLM INT8 quantization with 8-bit weights and 8-bit activations)
\end{itemize}

While quantized models enable faster and more efficient inference, users should be aware that they may exhibit a decrease in response quality compared to the full-precision versions.

In particular, the GGUF variants have been further optimized through a two-step process designed to preserve model quality after quantization:
\begin{itemize}
\item To minimize the degradation in output quality, we applied a post-training calibration procedure based on an importance matrix. The importance matrix assigns weights to model activations according to their contribution to the model's output, as estimated on a specially constructed bilingual Polish-English calibration dataset. This dataset was curated to reflect the target linguistic and task distributions. During calibration, layers or parameters deemed more important are preserved more precisely during quantization, leading to improved output consistency and better preservation of model capabilities, especially in multilingual settings.
 
\item Although the main model weights were quantized to low-bit formats (e.g., INT4), the embedding layers and the final output projection layer were retained in FP16 precision. This selective retention helps maintain the expressiveness of the input and output spaces, mitigating quality loss associated with aggressive quantization.
\end{itemize}

This calibration and quantization strategy improves the performance of GGUF models, making them suitable for deployment on resource-constrained devices while minimizing the loss in response quality.

\section{Evaluation}

We conducted a comprehensive evaluation of the Bielik 11B v2 models across multiple benchmarks to assess their performance in various linguistic tasks and languages. This section presents our evaluation results, comparing Bielik models against three key reference groups:

\begin{enumerate}
    \item Leading models across all parameter sizes, including Qwen \citep{qwen2,qwen2.5}, Llama \citep{grattafiori2024llama3herdmodels}, Mistral \citep{jiang2023mistral7b}, Aya 23 \citep{aryabumi2024aya}, and Gemma \citep{gemma_2024,gemma_2025}
    
    \item Models with similar parameter counts to Bielik, such as EuroLLM \citep{martins2024eurollmmultilinguallanguagemodels}, Teuken \citep{ali2024teuken7bbaseteuken7binstructeuropean}, Salamandra \citep{gonzalezagirre2025salamandratechnicalreport}, and Occiglot \citep{avramidis-etal-2024-occiglot}
    
    \item Other recent Polish language models including PLLuM \citep{pllum2025} and Qra
\end{enumerate}

For clarity, the tables presented throughout this section contain selected subsets of models from each leaderboard, chosen to provide meaningful comparisons. Complete and up-to-date results can be found on the respective leaderboards' online platforms.

The models were evaluated on the following benchmarks:

\begin{itemize}
    \item \href{https://huggingface.co/spaces/speakleash/open_pl_llm_leaderboard}{Open PL LLM Leaderboard} \textbf{(Polish)}
    \item \href{https://huggingface.co/spaces/speakleash/mt-bench-pl}{Polish MT-Bench} \textbf{(Polish)}
    \item \href{https://huggingface.co/spaces/speakleash/polish_eq-bench}{Polish EQ-Bench} \textbf{(Polish)}
    \item \href{https://huggingface.co/spaces/speakleash/cptu_bench}{CPTUB Leaderboard} \textbf{(Polish)}
    \item \href{https://huggingface.co/spaces/speakleash/polish_medical_leaderboard}{Polish Medical Leaderboard} \textbf{(Polish)}
    \item \href{https://huggingface.co/spaces/sdadas/plcc}{Polish Linguistic and Cultural Competency Benchmark (PLCC)} \textbf{(Polish)}
    \item \href{https://huggingface.co/spaces/amu-cai/LLMZSZL_Leaderboard}{LLMzSzŁ (LLMs Behind the School Desk)} \textbf{(Polish)}
    \item \href{https://huggingface.co/spaces/speakleash/european_leaderboard_bielik}{European LLM Leaderboard}
    \item \href{https://euroeval.com/leaderboards/Multilingual/european/}{EuroEval}
    \item \href{https://huggingface.co/spaces/open-llm-leaderboard-old/open_llm_leaderboard}{Open LLM Leaderboard}
    \item \href{https://huggingface.co/spaces/open-llm-leaderboard/open_llm_leaderboard/}{Open LLM Leaderboard v2}
    \item \href{https://mixeval.github.io/}{MixEval}
    \item \href{https://gorilla.cs.berkeley.edu/leaderboard.html}{Berkeley Function-Calling Leaderboard}
    \item \href{https://huggingface.co/spaces/speakleash/european_leaderboard_bielik}{FLORES200 Translation Benchmark} (see Appendix \ref{tab:flores-translation})
    \item \href{https://huggingface.co/spaces/CZLC/BenCzechMark}{BenCzechMark} (see Appendix \ref{app:benczechmark})
    \item \href{https://huggingface.co/spaces/eduagarcia/open_pt_llm_leaderboard}{Portuguese Benchmark (Open PT LLM Leaderboard)} (see Appendix \ref{app:portuguese})
\end{itemize}

It is important to note that Bielik 11B v2.5-Instruct is being released concurrently with this article. As such, comprehensive benchmark results for this latest version are not yet available across all evaluation frameworks mentioned above. The results presented in this section include data for Bielik 11B v2.5-Instruct where available, while predominately showcasing performance metrics for earlier v2 family models. Complete benchmark results for Bielik 11B v2.5-Instruct will be updated as evaluations are completed.

\subsection{Open PL LLM Leaderboard} \label{Open-PL-LLM-Leaderboard}

The Open PL LLM Leaderboard, based on the Open LLM Leaderboard v1 \citep{open-llm-leaderboard-v1}, evaluates models on various NLP tasks, including: sentiment analysis, categorization, short answer question answering, and text classification, but does not test their conversational capabilities \citep{open-pl-llm-leaderboard,ociepa2024bielik7bv01polish}. The leaderboard utilizes the lm-evaluation-harness framework for model evaluation \citep{eval-harness}.

\paragraph{Tasks:}
\begin{itemize}
    \item \textbf{polemo2:} Sentiment analysis of online consumer reviews across four domains (medicine, hotels, products, university) with four-class labeling (positive, negative, neutral, ambiguous) \citep{kocon-etal-2019-multi}; metric: accuracy.
    \item \textbf{klej-ner:} Named entity recognition in sentences containing single-type entities, classifying into six categories (no entity, place, person, organization, time, geographical name) \citep{rybak-etal-2020-klej}; metric: accuracy.
    \item \textbf{8tags:} Topic classification of social media headlines into eight categories (film, history, food, medicine, motorization, work, sport, technology) \citep{dadas-etal-2020-evaluation}; metric: accuracy.
    \item \textbf{belebele:} Machine reading comprehension for question answering \citep{bandarkar-etal-2024-belebele}; metric: accuracy.
    \item \textbf{dyk:} Question answering based on human-annotated pairs from Wikipedia's "Did You Know" section \citep{marcinczuk2013open}; metric: binary F1.
    \item \textbf{ppc:} Text similarity assessment using manually labeled sentence pairs (exact paraphrases, close paraphrases, non-paraphrases) \citep{9945218}; metric: accuracy.
    \item \textbf{psc:} Summarization of news articles \citep{ogro:kop:14:lrec}; metric: binary F1.
    \item \textbf{cbd:} Text classification for cyberbullying and hate-speech detection \citep{ptaszynski2023expert}; metric: macro F1.
    \item \textbf{polqa:} Open-domain question answering from the "Jeden z dziesięciu" TV show, with and without context (abstractive QA/RAG) \citep{rybak-etal-2024-polqa-polish}; metric: accuracy, levenshtein.
    \item \textbf{poquad:} Context-based extractive question answering (QA/RAG) \citep{tuora2023poquad}; metric: levenshtein.
    \item \textbf{eqbench:} emotional intelligence benchmark \citep{paech2024eqbenchemotionalintelligencebenchmark}; metric: custom.
\end{itemize}

Most of the tasks are multiple-choice tests, which means that the model chooses the correct answer from a set of options.
They are implemented as two types of tests:
\begin{itemize}
    \item \textbf{Loglikelihood:} We choose the highest probability token from the given set, e.g., ABCD. These tests are suitable for base models.
    \item \textbf{Generate:} Model generates answer freely.
\end{itemize}

All tasks are evaluated in both 0-shot and 5-shot settings, with the average score across all tasks normalized by baseline scores.

It is important to note that PLLuM models are not included in this leaderboard, as they were trained on training portions of the tasks used in the benchmark (except for the Belebele and EQ-Bench tasks), unlike all other models present on the leaderboard, to the best of our knowledge. The authors of the datasets used in the benchmark are primarily PLLuM consortium members.

\begin{table*}[t]
\centering
\small
\begin{tabular}{lrr}
\toprule
\textbf{Model} & \textbf{Parameters (B)} & \textbf{Average} \\
\midrule
Qwen2.5-72B & 72.7 & 67.38 \\
Qwen2.5-32B & 32.8 & 66.73 \\
Qwen-72B & 72.7 & 66.02 \\
Qwen2.5-14B & 14.8 & 62.71 \\
Meta-Llama-3-70B & 70.6 & 62.07 \\
Qwen1.5-72B & 72.7 & 61.11 \\
Meta-Llama-3.1-70B & 70.6 & 60.87 \\
Mixtral-8x22B-v0.1 & 141.0 & 60.75 \\
Mistral-Small-24B-Base-2501 & 24.0 & 59.90 \\
Qwen1.5-32B & 32.8 & 58.71 \\
\textbf{Bielik-11B-v2} & \textbf{11.2} & \textbf{58.14} \\
Qwen2.5-7B & 7.0 & 53.35 \\
EuroLLM-9B & 9.2 & 50.03 \\
Qwen-7B & 7.0 & 49.39 \\
SOLAR-10.7B-v1.0 & 10.7 & 47.54 \\
Mistral-Nemo-Base-2407 & 12.2 & 47.28 \\
internlm2-20b & 20.0 & 47.15 \\
Meta-Llama-3.1-8B & 8.0 & 43.77 \\
Meta-Llama-3-8B & 8.0 & 43.30 \\
Qwen1.5-72B & 72.3 & 39.51 \\
Mistral-7B-v0.3 & 7.0 & 38.88 \\
Mistral-7B-v0.2 & 7.0 & 38.81 \\
Qwen1.5-7B & 7.0 & 37.92 \\
\textbf{Bielik-7B-v0.1} & \textbf{7.2} & \textbf{34.34} \\
Qra-13b & 13.0 & 33.90 \\
Llama-3.2-3B & 3.0 & 31.89 \\
Qra-7b & 7.0 & 16.60 \\
\bottomrule
\end{tabular}
\caption{Open PL LLM Leaderboard results for base models (5-shot evaluation)}
\label{tab:open-pl-llm-base}
\end{table*}

As shown in Table \ref{tab:open-pl-llm-base}, Bielik-11B-v2 achieved an impressive average score of 58.14, making it competitive with much larger models. It obtains similar scores to Qwen1.5-32B (58.71) despite having significantly fewer parameters, and shows a substantial improvement over its predecessor, Bielik-7B-v0.1, which scored 34.34.

\begin{table*}[t]
\centering
\small
\begin{tabular}{lrr}
\toprule
\textbf{Model} & \textbf{Parameters (B)} & \textbf{Average} \\
\midrule
Mistral-Large-Instruct-2411 & 123.0 & 69.84 \\
Meta-Llama-3.1-405B-Instruct-FP8 & 405.0 & 69.44 \\
Mistral-Large-Instruct-2407 & 123.0 & 69.11 \\
Qwen2.5-72B-Instruct & 72.7 & 67.92 \\
QwQ-32B-Preview & 32.8 & 67.01 \\
Llama-3.3-70B-Instruct & 70.6 & 66.40 \\
Qwen2-72B-Instruct & 72.7 & 65.87 \\
\textbf{Bielik-11B-v2.3-Instruct} & \textbf{11.2} & \textbf{65.71} \\
\textbf{Bielik-11B-v2.2-Instruct} & \textbf{11.2} & \textbf{65.57} \\
Meta-Llama-3.1-70B-Instruct & 70.6 & 65.49 \\
\textbf{Bielik-11B-v2.1-Instruct} & \textbf{11.2} & \textbf{65.45} \\
Mixtral-8x22B-Instruct-v0.1 & 141.0 & 65.23 \\
\textbf{Bielik-11B-v2.0-Instruct} & \textbf{11.2} & \textbf{64.98} \\
Meta-Llama-3-70B-Instruct & 70.6 & 64.45 \\
Qwen3-32B & 32.8 & 64.24 \\
Llama-4-Scout-17B-16E-Instruct & 109.0 & 64.21 \\
\textbf{Bielik-11B-v2.5-Instruct} & \textbf{11.2} & \textbf{63.95} \\
Mistral-Small-24B-Instruct-2501 & 24.0 & 62.97 \\
phi-4 & 14.7 & 62.57 \\
Qwen3-14B & 14.8 & 62.24 \\
Mistral-Small-Instruct-2409 & 22.2 & 61.41 \\
Qwen2.5-32B-Instruct & 32.8 & 61.21 \\
Qwen2.5-14B-Instruct & 14.8 & 59.91 \\
aya-23-35B & 35.0 & 56.37 \\
Qwen3-8B & 8.2 & 55.78 \\
Qwen3-4B & 4.0 & 55.49 \\
Mistral-Nemo-Instruct-2407 & 12.2 & 55.27 \\
Qwen2.5-7B-Instruct & 7.6 & 54.93 \\
Mistral-7B-Instruct-v0.3 & 7.2 & 47.74 \\
Mistral-7B-Instruct-v0.2 & 7.2 & 45.95 \\
\textbf{Bielik-7B-Instruct-v0.1} & \textbf{7.2} & \textbf{44.70} \\
Qwen2.5-3B-Instruct & 3.1 & 41.23 \\
Qwen3-1.7B & 2.0 & 38.34 \\
Mistral-7B-Instruct-v0.1 &	7.0 &	33.11 \\
Qwen2.5-1.5B-Instruct & 1.5 & 31.89 \\
\bottomrule
\end{tabular}
\caption{Open PL LLM Leaderboard results for instruction-tuned models (5-shot evaluation)}
\label{tab:open-pl-llm-instruct}
\end{table*}

The instruction-tuned versions show further performance improvements. The latest results from the Open PL LLM Leaderboard (Table~\ref{tab:open-pl-llm-instruct}) highlight the outstanding performance of the Bielik models. The Bielik-11B-v2.3-Instruct achieves a remarkable score of 65.71, placing it among the top performers and outperforming many models with significantly more parameters.

Several key observations emerge from these results:

\begin{enumerate}
    \item The entire Bielik-11B-v2.x family performs exceptionally well, with scores ranging from 64.98 to 65.71, placing them ahead of many larger models including Meta-Llama-3-70B-Instruct (64.45).

    \item The Bielik-11B-v2.3-Instruct serves as the flagship model, with the Bielik-11B-v2.2-Instruct version following closely at 65.57.
    
    \item The latest Bielik-11B-v2.5-Instruct model achieves a strong score of 63.95. While slightly lower than its v2.3 predecessor, this version incorporates enhanced function-calling capabilities and other architectural improvements that trade off some benchmark performance for improved practical functionality.

    \item When considering parameter efficiency, Bielik models are particularly impressive, outperforming many models with 2-6 times more parameters.

    \item The evolution from Bielik-7B-Instruct-v0.1 (44.70) to the v2 series shows a dramatic improvement of more than 20 percentage points.
\end{enumerate}

\subsubsection{Quantization Performance}

Model quantization is essential for deploying large language models efficiently on consumer hardware. The Open PL LLM benchmark provides valuable insights into how different quantization methods affect Bielik's performance, showing its practical utility in resource-constrained environments.

\begin{table*}[t]
\centering
\small
\begin{tabular}{lrr}
\toprule
\textbf{Model} & \textbf{Params (B)} & \textbf{Average (\%)} \\
\midrule
Bielik-11B-v2.3-Instruct.Q8\_0.gguf & 11.2 & 65.76 \\
\textbf{Bielik-11B-v2.3-Instruct} & \textbf{11.2} & \textbf{65.71} \\
Bielik-11B-v2.3-Instruct.Q6\_K.gguf & 11.2 & 65.26 \\
Bielik-11B-v2.3-Instruct.IQ3\_XXS.gguf.IQ & 11.2 & 64.89 \\
Bielik-11B-v2.3-Instruct.Q4\_K\_M.gguf & 11.2 & 64.76 \\
Bielik-11B-v2.3-Instruct.Q4\_K\_M.gguf.IQ & 11.2 & 64.71 \\
Bielik-11B-v2.3-Instruct.IQ2\_XXS.gguf.IQ & 11.2 & 61.34 \\
Mistral-Nemo-Instruct-2407 & 12.2 & 55.27 \\
Bielik-11B-v2.3-Instruct.IQ1\_M.gguf.IQ & 11.2 & 52.09 \\
Mistral-7B-Instruct-v0.3 & 7.2 & 47.74 \\
Bielik-7B-Instruct-v0.1 & 7.2 & 44.70 \\
\bottomrule
\end{tabular}
\caption{Open PL LLM benchmark results for Bielik-11B-v2.3-Instruct across different quantization methods, compared to other models. Higher scores are better.}
\label{tab:quantization-results}
\end{table*}

\paragraph{Resilience to quantization:} Bielik-11B-v2.3-Instruct demonstrates remarkable resilience across different quantization methods (as shown in Table~\ref{tab:quantization-results}):
\begin{itemize}
    \item The Q8\_0 quantized version (65.76\%) actually slightly outperforms the original full-precision model (65.71\%)
    \item More aggressive quantization with Q6\_K (65.26\%), IQ3\_XXS (64.89\%), and Q4\_K\_M (64.76\%) maintains over 98\% of the original performance
    \item Even at extreme compression with IQ2\_XXS (61.34\%), Bielik preserves 93\% of its original capability
    \item The most aggressive IQ1\_M quantization (52.09\%) still outperforms both Mistral-7B-Instruct-v0.3 (47.74\%) and Bielik-7B-Instruct-v0.1 (44.70\%)
\end{itemize}

\paragraph{Practical implications:} These results have significant implications for real-world applications:
\begin{itemize}
    \item Bielik can be effectively deployed on consumer-grade hardware with minimal performance degradation
    \item The Q4\_K\_M quantization offers an excellent balance of model size reduction and performance preservation
    \item Even highly compressed versions remain competitive with larger models, enabling deployment in memory-constrained environments
    \item The model architecture appears inherently robust to quantization effects, suggesting effective weight distribution during training
\end{itemize}

This quantization analysis demonstrates Bielik's practical utility beyond benchmark scores, showing that its strong performance can be maintained even under significant compression. This makes the model particularly valuable for applications where deployment efficiency is critical without sacrificing Polish language understanding capabilities.

\subsection{Polish MT-Bench}

MT-bench \cite{zheng2023judging} is a tool designed to test the ability of language models (LLMs) to conduct two-step conversations and follow instructions. It covers typical use cases and focuses on challenging questions to differentiate the capabilities of various models. Eight main categories of user queries were identified, which were used to construct MT-bench:

\begin{itemize}
    \item writing
    \item role-playing
    \item information extraction
    \item reasoning
    \item mathematics
    \item coding
    \item knowledge / hard sciences / stem
    \item knowledge / humanities / social sciences
\end{itemize}

The evaluation of responses is performed by a metamodel. In the case of MT-Bench, this is the GPT-4 model. By using a metamodel, we can verify responses from open-ended questions, e.g., write an article about hybrid cars. The model evaluates the content of the response, the quality of facts used, creativity, etc.

The Polish MT-Bench \cite{rkinas2024mt-bench-pl} has been completely polonized. Each task was first machine-translated and then verified. Additionally, we introduced Polish accents, e.g., instead of describing a vacation in Hawaii, we suggested the location - Mazury (Masuria). In our language version, many changes were introduced to transfer the test into Polish linguistic realities.

\begin{table*}[t]
\centering
\small
\begin{tabular}{lrccccccccc}
\toprule
\textbf{Model} & \textbf{Params (B)} & \textbf{Score} & \textbf{Coding} & \textbf{Extract.} & \textbf{Human.} & \textbf{Math} & \textbf{Reason.} & \textbf{Role-play} & \textbf{STEM} & \textbf{Writing} \\
\midrule
gemma-3-27b-it & 27.4 & 9.28 & 8.10 & 9.90 & 10.00 & 8.25 & 8.40 & 9.95 & 9.95 & 9.70 \\
Mistral-3.1-24B & 24.0 & 9.18 & 8.30 & 9.80 & 10.00 & 7.85 & 9.00 & 9.40 & 9.90 & 9.15 \\
phi-4 & 14.7 & 9.07 & 7.60 & 9.30 & 9.95 & 7.70 & 9.55 & 9.20 & 10.00 & 9.25 \\
gemma-3-12b-it & 12.0 & 8.97 & 8.25 & 9.55 & 10.00 & 7.45 & 7.75 & 9.45 & 10.00 & 9.30 \\
Qwen2.5-32B-Instruct & 32.8 & 8.86 & 7.95 & 9.90 & 9.65 & 7.60 & 9.10 & 8.30 & 9.70 & 8.65 \\
Qwen2-72B-Instruct & 72.7 & 8.78 & 7.80 & 9.80 & 9.75 & 6.50 & 8.85 & 9.20 & 9.55 & 8.75 \\
Mistral-Small-2501 & 24.0 & 8.72 & 7.95 & 9.90 & 9.70 & 7.83 & 7.90 & 9.05 & 9.50 & 7.95 \\
Mistral-Large-2407 & 123.0 & 8.66 & 6.75 & 9.90 & 9.40 & 7.80 & 8.70 & 8.70 & 9.35 & 8.70 \\
aya-expanse-32b & 32.0 & 8.62 & 5.75 & 8.40 & 10.00 & 6.60 & 8.95 & 9.70 & 9.95 & 9.60 \\
gemma-2-27b-it & 27.4 & 8.62 & 7.45 & 9.60 & 10.00 & 7.80 & 6.85 & 8.70 & 9.80 & 8.75 \\
Mistral-Small-2409 & 22.2 & 8.56 & 7.10 & 9.15 & 10.00 & 7.00 & 7.90 & 8.90 & 9.65 & 8.80 \\
\textbf{Bielik-11B-v2.3-Instruct} & \textbf{11.2} & \textbf{8.56} & \textbf{6.25} & \textbf{9.43} & \textbf{9.50} & \textbf{7.70} & \textbf{8.35} & \textbf{8.75} & \textbf{8.98} & \textbf{9.50} \\
Qwen2.5-14B-Instruct & 14.8 & 8.33 & 6.70 & 9.25 & 9.18 & 8.10 & 7.55 & 8.50 & 9.60 & 7.75 \\
gemma-3-4b-it & 4.0 & 8.22 & 5.40 & 8.40 & 9.90 & 7.40 & 6.25 & 9.45 & 9.65 & 9.30 \\
Meta-Llama-3.1-405B & 405.0 & 8.17 & 7.25 & 9.85 & 9.65 & 6.25 & 5.80 & 8.70 & 8.65 & 9.20 \\
Meta-Llama-3.1-70B & 70.6 & 8.15 & 6.25 & 9.85 & 9.50 & 6.00 & 6.15 & 8.80 & 9.55 & 9.10 \\
\textbf{Bielik-11B-v2.2-Instruct} & \textbf{11.2} & \textbf{8.12} & \textbf{5.05} & \textbf{9.30} & \textbf{9.40} & \textbf{6.45} & \textbf{6.90} & \textbf{9.03} & \textbf{9.45} & \textbf{9.35} \\
\textbf{Bielik-11B-v2.1-Instruct} & \textbf{11.2} & \textbf{8.00} & \textbf{5.40} & \textbf{9.13} & \textbf{9.20} & \textbf{6.15} & \textbf{6.25} & \textbf{9.45} & \textbf{8.90} & \textbf{9.50} \\
aya-expanse-8b & 8.0 & 7.76 & 4.90 & 8.05 & 9.65 & 4.35 & 6.85 & 9.25 & 9.75 & 9.30 \\
gpt-3.5-turbo & Unknown & 7.72 & 6.00 & 8.15 & 9.75 & 6.85 & 5.20 & 8.65 & 9.25 & 9.10 \\
Mixtral-8x7b & 46.7 & 7.64 & 5.20 & 8.15 & 9.45 & 5.65 & 5.80 & 8.95 & 8.55 & 9.35 \\
\textbf{Bielik-11B-v2.0-Instruct} & \textbf{11.2} & \textbf{7.56} & \textbf{5.60} & \textbf{8.65} & \textbf{9.43} & \textbf{5.50} & \textbf{6.05} & \textbf{7.75} & \textbf{8.78} & \textbf{8.75} \\
Mistral-Nemo-2407 & 12.2 & 7.37 & 5.85 & 8.95 & 9.50 & 6.70 & 5.80 & 7.45 & 8.30 & 6.40 \\
Llama-PLLuM-70B-chat & 70.6 & 6.75 & 4.80 & 9.45 & 8.80 & 2.90 & 5.20 & 6.60 & 8.20 & 8.05 \\
openchat-3.5-0106-gemma & 7.0 & 6.51 & 5.35 & 6.90 & 8.80 & 4.55 & 5.40 & 7.98 & 8.48 & 7.05 \\
PLLuM-12B-nc-chat$^{\dagger}$ & 12.2 & 6.47 & 4.55 & 7.20 & 9.50 & 2.30 & 4.80 & 6.75 & 9.10 & 7.55 \\
PLLuM-8x7B-nc-chat$^{\dagger}$ & 46.7 & 6.43 & 4.10 & 8.40 & 7.48 & 3.35 & 4.95 & 6.90 & 8.90 & 7.40 \\
PLLuM-8x7B-chat & 46.7 & 6.30 & 4.55 & 8.00 & 8.60 & 3.45 & 4.90 & 6.25 & 8.20 & 7.10 \\
Llama-PLLuM-8B-chat & 8.0 & 6.05 & 3.65 & 6.30 & 9.50 & 2.75 & 5.35 & 6.15 & 7.50 & 7.20 \\
PLLuM-12B-chat & 12.2 & 5.81 & 3.05 & 6.55 & 9.30 & 2.65 & 3.90 & 5.00 & 8.00 & 8.00 \\
\textbf{Bielik-7B-Instruct-v0.1} & \textbf{7.2} & \textbf{5.40} & \textbf{3.00} & \textbf{4.35} & \textbf{8.48} & \textbf{4.10} & \textbf{6.15} & \textbf{7.83} & \textbf{6.90} & \textbf{7.85} \\
dolphin-2.9.1-llama-3-8b & 8.0 & 5.24 & 4.60 & 6.15 & 8.80 & 4.80 & 3.30 & 7.40 & 6.35 & 5.50 \\
Polka-Mistral-7B-SFT & 7.0 & 4.43 & 2.95 & 5.25 & 5.60 & 2.95 & 2.45 & 4.90 & 6.80 & 5.25 \\
trurl-2-7b & 7.0 & 2.75 & 1.80 & 3.50 & 3.95 & 1.70 & 2.05 & 3.30 & 2.65 & 3.15 \\
\bottomrule
\multicolumn{11}{l}{$^{\dagger}$Models with a non-commercial license.} \\
\end{tabular}
\caption{Polish MT-Bench results for various models across all evaluation categories.}
\label{tab:pl-mt-bench-scores}
\end{table*}

The latest results from the Polish MT-Bench (Table~\ref{tab:pl-mt-bench-scores}) reveal several key insights about the Bielik models' performance:

\paragraph{Top-tier performance:} Bielik-11B-v2.3-Instruct achieves a remarkable score of 8.56, placing it in the top tier of models alongside much larger architectures. It matches the performance of Mistral-Small-Instruct-2409 (22.2B) and outperforms larger models including Meta-Llama-3.1-405B-Instruct (405B) and Meta-Llama-3.1-70B-Instruct (70B).

\paragraph{Efficiency advantage:} When considering parameter count efficiency, Bielik-11B-v2.3-Instruct demonstrates exceptional performance. It achieves comparable or better results than models with 2-6 times more parameters, including Qwen2.5-14B-Instruct (14B) and aya-expanse-32b (32B).

\paragraph{Category strengths:} Examining performance across categories, Bielik-11B-v2.3-Instruct shows particular strength in mathematics (7.70), reasoning (8.35), and writing (9.50). The writing score is especially noteworthy, matching gemma-3-27b-it despite the latter having over twice the parameters. This balanced performance across all categories indicates versatility rather than specialization in particular areas.

\paragraph{Comparison with other Polish models:} The performance gap between Bielik and other dedicated Polish language models is substantial. Bielik-11B-v2.3-Instruct outperforms the best PLLuM model (Llama-PLLuM-70B-chat, 6.75) by 1.81 points despite having less than 16\% of its parameters. Even more striking, it surpasses PLLuM-12B-nc-chat (6.47) by 2.09 points despite similar parameter counts.

\paragraph{Evolutionary improvements:} The progression from Bielik-7B-Instruct-v0.1 (5.40) to Bielik-11B-v2.3-Instruct (8.56) demonstrates a dramatic 3.16-point improvement. Even within the v2 series, each iteration shows clear advancements, with v2.3 improving 0.44 points over v2.2, 0.56 points over v2.1, and 1.00 points over v2.0.

These results highlight Bielik-11B-v2.3-Instruct's position as a state-of-the-art model for Polish language processing, delivering performance comparable to much larger models while maintaining efficiency in parameter count.

\subsection{Polish EQ-Bench}

The Polish Emotional Intelligence Benchmark, which is a localized Polish version of the original EQ-Bench \citep{paech2024eqbenchemotionalintelligencebenchmark}, evaluates the emotional intelligence capabilities of language models across various dimensions of emotional understanding and response.

\begin{table*}[t]
\centering
\small
\begin{tabular}{lrc}
\toprule
\textbf{Model} & \textbf{Parameters (B)} & \textbf{Score} \\
\midrule
Mistral-Large-Instruct-2407$^{\dagger}$ & 123.0 & 78.07 \\
Mistral-Large-Instruct-2411$^{\dagger}$ & 123.0 & 77.29 \\
Meta-Llama-3.1-405B-Instruct-FP8 & 405.0 & 77.23 \\
gpt-4o-2024-08-06 & Unknown & 75.15 \\
gpt-4-turbo-2024-04-09 & Unknown & 74.59 \\
Mistral-Small-Instruct-2409 & 22.2 & 72.85 \\
Llama-PLLuM-70B-chat & 70.6 & 72.56 \\
Meta-Llama-3.1-70B-Instruct & 70.6 & 72.53 \\
\textbf{Bielik-11B-v2.5-Instruct} & \textbf{11.2} & \textbf{72.00} \\
Qwen2-72B-Instruct & 72.7 & 71.23 \\
Meta-Llama-3-70B-Instruct & 70.6 & 71.21 \\
gpt-4o-mini-2024-07-18 & Unknown & 71.15 \\
Qwen2.5-32B-Instruct & 32.8 & 71.15 \\
\textbf{Bielik-11B-v2.3-Instruct} & \textbf{11.2} & \textbf{70.86} \\
Llama-3.3-70B-Instruct & 70.6 & 70.73 \\
Llama-PLLuM-70B-instruct & 70.6 & 69.99 \\
WizardLM-2-8x22B & 141.0 & 69.56 \\
Qwen2.5-14B-Instruct & 14.8 & 69.17 \\
\textbf{Bielik-11B-v2.2-Instruct} & \textbf{11.2} & \textbf{69.05} \\
\textbf{Bielik-11B-v2.0-Instruct} & \textbf{11.2} & \textbf{68.24} \\
glm-4-9b-chat & 9.0 & 61.79 \\
Mistral-Nemo-Instruct-2407 & 12.2 & 61.76 \\
\textbf{Bielik-11B-v2.1-Instruct} & \textbf{11.2} & \textbf{60.07} \\
PLLuM-12B-chat & 12.2 & 52.26 \\
PLLuM-8x7B-nc-chat$^{\dagger}$ & 46.7 & 47.29 \\
Llama-PLLuM-8B-chat & 8.0 & 46.20 \\
PLLuM-8x7B-chat & 46.7 & 45.22 \\
PLLuM-8x7B-nc-instruct$^{\dagger}$ & 46.7 & 41.75 \\
PLLuM-8x7B-instruct & 46.7 & 39.55 \\
PLLuM-12B-instruct & 12.2 & 36.21 \\
PLLuM-12B-nc-chat$^{\dagger}$ & 12.2 & 35.41 \\
Llama-PLLuM-8B-instruct & 8.0 & 31.59 \\
PLLuM-12B-nc-instruct$^{\dagger}$ & 12.2 & 13.11 \\
\bottomrule
\multicolumn{3}{l}{$^{\dagger}$Models with a non-commercial license.} \\
\end{tabular}
\caption{Polish EQ-Bench results for various models.}
\label{tab:pl-eq-bench}
\end{table*}

Bielik-11B-v2.3-Instruct demonstrates exceptional emotional intelligence capabilities with a score of 70.86 (as shown in Table~\ref{tab:pl-eq-bench}), outperforming many larger models including WizardLM-2-8x22B (141B parameters). This places it just 1.67 points behind Qwen2.5-32B-Instruct despite having only a third of the parameters. 

The newest addition to the family, Bielik-11B-v2.5-Instruct, further improves on these results with a score of 72.00, surpassing both Qwen2-72B-Instruct (71.23) and Meta-Llama-3-70B-Instruct (71.21) despite having just a fraction of their parameters. This positions it less than 0.53 points behind Meta-Llama-3.1-70B-Instruct (72.53), showcasing the effectiveness of the latest training methodology.

In the comparison with other Polish language models Bielik-11B-v2.3-Instruct outperformed most of PLLuM model variants except the largest 70B models. It surpasses:
\begin{itemize}
    \item PLLuM-12B-chat by 18.60 points (70.86 vs 52.26)
    \item PLLuM-8x7B-nc-chat by 23.57 points (70.86 vs 47.29) (MoE architecture with 46.7B effective parameters)
    \item Smaller PLLuM variants by 25-57 points
\end{itemize}

Among the Bielik variants, v2.3-Instruct demonstrates clear superiority over its predecessors, with a 1.81 point improvement over v2.2-Instruct, 2.62 points over v2.1, and a substantial 10.79 point advantage over v2.0. The gap between Bielik and the top-performing models (Mistral-Large and Llama-3.1-405B) is remarkably narrow considering the significant difference in model scale.

\subsection{Complex Polish Text Understanding Benchmark (CPTUB)}

The Complex Polish Text Understanding Benchmark (CPTUB) \citep{cptub-leaderboard} is specifically designed to evaluate language models' proficiency in interpreting complex Polish texts. Unlike traditional tasks that focus on explicit meaning, CPTUB assesses the models' capacity to understand implied meanings and handle cognitively challenging questions. The benchmark comprises two main components:

\begin{itemize}
    \item \textbf{Implicatures}: Evaluates a model's ability to interpret implied meanings, including sarcasm, idiomatic expressions, and phraseological compounds. This component tests sensitivity to nuanced, context-dependent inferences through three subtasks:
    \begin{itemize}
        \item \textbf{Sentiment}: Correctly identifying the emotional tone beyond literal expressions
        \item \textbf{Language understanding}: Interpreting the underlying intentions of text authors
        \item \textbf{Phraseology}: Recognizing and explaining fixed or semi-fixed expressions whose meanings cannot be inferred from their individual components
    \end{itemize}
    \item \textbf{Tricky Questions}: Assesses a model's capability to address challenging questions characterized by logical puzzles, semantic ambiguity, logical inconsistencies, absurdity, and humor. This component specifically targets the model's reasoning skills and ability to avoid hallucinations when faced with ambiguous or nonsensical queries.
\end{itemize}

\begin{table*}[t]
\centering
\small
\begin{tabular}{lrccccccc}
\toprule
\textbf{Model} & \textbf{Params (B)} & \textbf{Overall} & \textbf{Implicatures} & \textbf{Senti-} & \textbf{Language} & \textbf{Phrase-} & \textbf{Tricky} \\
& & \textbf{Average} & \textbf{Average} & \textbf{ment} & \textbf{Understanding} & \textbf{ology} & \textbf{Questions} \\
\midrule
DeepSeek-R1 & 685.0 & 4.14 & 4.14 & 4.49 & 4.35 & 3.60 & 4.12 \\
Mistral-Large-Instruct-2411$^{\dagger}$ & 123.0 & 4.00 & 4.10 & 4.33 & 3.98 & 3.99 & 3.72 \\
Qwen2.5-72B-Instruct & 72.7 & 3.95 & 3.99 & 4.08 & 3.97 & 3.93 & 3.81 \\
Mistral-Large-Instruct-2407$^{\dagger}$ & 123.0 & 3.93 & 4.03 & 4.23 & 4.00 & 3.86 & 3.65 \\
Llama-4-Maverick-17B-128E-Instruct-FP8 & 402.0 & 3.93 & 3.99 & 4.39 & 4.11 & 3.48 & 3.76 \\
gemma-3-27b-it & 27.4 & 3.81 & 3.90 & 3.88 & 3.79 & 4.03 & 3.53 \\
Meta-Llama-3-70B-Instruct & 70.6 & 3.78 & 3.81 & 4.13 & 3.82 & 3.47 & 3.71 \\
Qwen2.5-32B-Instruct & 32.8 & 3.75 & 3.80 & 3.81 & 3.57 & 4.04 & 3.59 \\
Llama-4-Scout-17B-16E-Instruct & 109.0 & 3.75 & 3.94 & 4.10 & 3.81 & 3.90 & 3.19 \\
\textbf{Bielik-11B-v2.3-Instruct} & \textbf{11.2} & \textbf{3.63} & \textbf{3.77} & \textbf{3.97} & \textbf{3.79} & \textbf{3.55} & \textbf{3.22} \\
\textbf{Bielik-11B-v2.1-Instruct} & \textbf{11.2} & \textbf{3.61} & \textbf{3.66} & \textbf{3.96} & \textbf{3.92} & \textbf{3.11} & \textbf{3.47} \\
Mixtral-8x22B-Instruct-v0.1 & 141.0 & 3.56 & 3.67 & 3.78 & 3.68 & 3.55 & 3.24 \\
Qwen2.5-14B-Instruct & 14.8 & 3.55 & 3.62 & 3.91 & 3.57 & 3.37 & 3.34 \\
Llama-PLLuM-70B-chat & 70.6 & 3.53 & 3.63 & 3.94 & 3.61 & 3.35 & 3.21 \\
\textbf{Bielik-11B-v2.5-Instruct} & \textbf{11.2} & \textbf{3.48} & \textbf{3.67} & \textbf{4.01} & \textbf{3.86} & \textbf{3.13} & \textbf{2.91} \\
\textbf{Bielik-11B-v2.2-Instruct} & \textbf{11.2} & \textbf{3.46} & \textbf{3.57} & \textbf{3.72} & \textbf{3.73} & \textbf{3.25} & \textbf{3.12} \\
Llama-PLLuM-70B-instruct & 70.6 & 3.33 & 3.56 & 3.78 & 3.63 & 3.26 & 2.63 \\
phi-4 & 14.7 & 3.30 & 3.50 & 3.72 & 3.54 & 3.24 & 2.72 \\
\textbf{Bielik-11B-v2.0-Instruct} & \textbf{11.2} & \textbf{3.26} & \textbf{3.61} & \textbf{3.97} & \textbf{3.75} & \textbf{3.13} & \textbf{2.20} \\
PLLuM-12B-nc-chat$^{\dagger}$ & 12.2 & 3.15 & 3.33 & 3.22 & 3.23 & 3.54 & 2.62 \\
PLLuM-12B-chat & 12.2 & 3.14 & 3.32 & 3.32 & 3.21 & 3.43 & 2.59 \\
PLLuM-8x7B-nc-instruct$^{\dagger}$ & 46.7 & 3.11 & 3.56 & 3.88 & 3.59 & 3.22 & 1.76 \\
PLLuM-12B-instruct & 12.2 & 3.09 & 3.49 & 3.71 & 3.17 & 3.59 & 1.90 \\
Qwen2.5-7B-Instruct & 7.62 & 3.07 & 3.23 & 3.56 & 3.03 & 3.10 & 2.58 \\
PLLuM-8x7B-nc-chat$^{\dagger}$ & 46.7 & 3.03 & 3.44 & 3.76 & 3.48 & 3.08 & 1.80 \\
Meta-Llama-3.1-8B-Instruct & 8.0 & 3.01 & 3.31 & 3.97 & 3.38 & 2.58 & 2.11 \\
PLLuM-8x7B-instruct & 46.7 & 3.01 & 3.51 & 3.59 & 3.47 & 3.46 & 1.51 \\
PLLuM-8x7B-chat & 46.7 & 3.01 & 3.41 & 3.44 & 3.45 & 3.35 & 1.78 \\
Meta-Llama-3-8B-Instruct & 8.0 & 3.00 & 3.17 & 3.33 & 3.15 & 3.04 & 2.48 \\
Llama-PLLuM-8B-chat & 8.0 & 2.92 & 3.14 & 3.13 & 2.93 & 3.36 & 2.25 \\
\textbf{Bielik-7B-Instruct-v0.1} & \textbf{7.2} & \textbf{2.88} & \textbf{3.13} & \textbf{3.59} & \textbf{3.48} & \textbf{2.32} & \textbf{2.16} \\
Llama-PLLuM-8B-instruct & 8.0 & 2.82 & 3.20 & 3.24 & 2.90 & 3.46 & 1.66 \\
\bottomrule
\multicolumn{8}{l}{$^{\dagger}$Models with a non-commercial license.} \\
\end{tabular}
\caption{Complex Polish Text Understanding Benchmark (CPTUB) results across different evaluation categories}
\label{tab:cptub}
\end{table*}

As shown in Table \ref{tab:cptub}, Bielik models demonstrate strong performance on this challenging benchmark, particularly in understanding implied meanings in Polish text. The best-performing variant, Bielik-11B-v2.3-Instruct, achieves an overall score of 3.63, placing it ahead of many larger models, including Mixtral-8x22B-Instruct-v0.1 (3.56).

Several observations can be made from these results:

\begin{enumerate}
    \item Bielik-11B-v2.3-Instruct and Bielik-11B-v2.1-Instruct show particularly strong performance in sentiment analysis (3.97 and 3.96) and language understanding (3.79 and 3.92), indicating robust capabilities in interpreting emotional cues and author intentions in complex Polish texts.
    
    \item The latest Bielik-11B-v2.5-Instruct achieves a score of 3.48, with notable strength in sentiment analysis (4.01) and language understanding (3.86). Its performance in tricky questions (2.91) is somewhat lower than previous versions, reflecting its optimization focus on other capabilities.
    
    \item The performance gap between Bielik models and similarly-sized models is notable. For instance, Bielik-11B-v2.3-Instruct (3.63) substantially outperforms phi-4 (3.30) despite having fewer parameters.
    
    \item When compared to other Polish language models, Bielik models show clear superiority over the PLLuM family. Bielik-11B-v2.3-Instruct (3.63) significantly outperforms even the largest PLLuM model, Llama-PLLuM-70B-chat (3.53), despite having only about 16\% of its parameters.
    
    \item In the challenging "Tricky Questions" category, which tests reasoning and hallucination avoidance, Bielik-11B-v2.1-Instruct achieves the best score among the Bielik family at 3.47, demonstrating its robustness against ambiguous or nonsensical queries.
\end{enumerate}

These results highlight the effectiveness of Bielik models in handling complex linguistic phenomena specific to the Polish language, suggesting that the training methodology has successfully captured nuanced aspects of Polish language understanding beyond simple translation or basic comprehension tasks.

\subsection{Polish Medical Leaderboard}

The Polish Medical Leaderboard evaluates language models on Polish Board Certification Examinations (Państwowy Egzamin Specjalizacyjny, PES) from years 2018-2022. This benchmark assesses models' medical knowledge and reasoning capabilities in a Polish-language medical context, using datasets from speakleash/PES-2018-2022, which is based on amu-cai/PES-2018-2022 \cite{pokrywka2024gpt4passes}.

\begin{table*}[t]
\centering
\small
\begin{tabular}{lrc}
\toprule
\textbf{Model} & \textbf{Parameters (B)} & \textbf{Average (\%)} \\
\midrule
Meta-Llama-3.1-405B-Instruct-FP8 & 405.0 & 69.20 \\
Mistral-Large-Instruct-2407$^{\dagger}$ & 123.0 & 64.28 \\
Qwen2.5-72B-Instruct & 72.7 & 63.89 \\
Meta-Llama-3.1-70B-Instruct & 70.6 & 61.75 \\
Qwen2-72B-Instruct & 72.7 & 61.35 \\
Meta-Llama-3-70B-Instruct & 70.6 & 57.51 \\
Qwen2.5-32B & 32.8 & 55.69 \\
Qwen2.5-32B-Instruct & 32.8 & 54.52 \\
Qwen2.5-14B-Instruct & 14.8 & 49.60 \\
\textbf{Bielik-11B-v2.5-Instruct} & \textbf{11.2} & \textbf{44.85} \\
GLM-4-9b-chat & 9.0 & 44.54 \\
Mistral-Small-Instruct-2409 & 22.2 & 43.60 \\
\textbf{Bielik-11B-v2.3-Instruct} & \textbf{11.2} & \textbf{43.26} \\
\textbf{Bielik-11B-v2.1-Instruct} & \textbf{11.2} & \textbf{43.16} \\
\textbf{Bielik-11B-v2.2-Instruct} & \textbf{11.2} & \textbf{43.05} \\
Qwen2.5-7B-Instruct & 7.6 & 42.69 \\
\textbf{Bielik-11B-v2.0-Instruct} & \textbf{11.2} & \textbf{41.53} \\
Meta-Llama-3.1-8B-Instruct & 8.0 & 40.60 \\
Mistral-Nemo-Instruct-2407 & 12.2 & 40.36 \\
\textbf{Bielik-11B-v2} & \textbf{11.2} & \textbf{39.98} \\
\bottomrule
\multicolumn{3}{l}{$^{\dagger}$Models with a non-commercial license.} \\
\end{tabular}
\caption{Polish Medical Leaderboard results (5-shot setting) showing model performance on Polish Board Certification Examinations.}
\label{tab:medical-leaderboard}
\end{table*}

\paragraph{Bielik's performance:} In the Polish Medical Leaderboard (Table~\ref{tab:medical-leaderboard}), Bielik-11B-v2.3-Instruct achieves a score of 43.26\%, demonstrating solid medical reasoning capabilities in Polish:
\begin{itemize}
    \item The latest Bielik-11B-v2.5-Instruct shows further improvement with a score of 44.85\%, marking the best performance among all Bielik variants
    \item The model performs competitively among models of similar size, outperforming Meta-Llama-3.1-8B-Instruct (40.60\%)
    \item The entire Bielik-11B-v2.x family shows consistent performance in the 41-45\% range
    \item Bielik models outperform Mistral-Nemo-Instruct models
\end{itemize}

\paragraph{Performance context:} The benchmark highlights the following insights about Bielik's medical capabilities:
\begin{itemize}
    \item Bielik demonstrates respectable medical reasoning without domain-specific training
    \item The model's performance points to strong cross-domain generalization from general Polish language understanding to specialized medical knowledge
    \item The gap between Bielik (43.26\%) and top-performing models like Meta-Llama-3.1-405B-Instruct (69.20\%) is expected given the significant difference in model scale (11.2B vs 405B parameters)
\end{itemize}

These results highlight Bielik's versatility across different knowledge domains, achieving competitive performance on specialized medical examinations despite not being specifically trained for medical applications. This versatility makes Bielik suitable for a wide range of practical applications in Polish language processing.

\subsection{Polish Linguistic and Cultural Competency Benchmark (PLCC)}

The Polish Linguistic and Cultural Competency Benchmark (PLCC) \citep{dadas2025evaluatingpolishlinguisticcultural} evaluates language models' knowledge of Polish cultural context, moving beyond traditional NLP tasks to assess cultural understanding. The benchmark consists of 600 manually crafted questions across six categories: history, geography, culture \& tradition, art \& entertainment, grammar, and vocabulary. 

PLCC questions assess knowledge of Polish cultural references, historical events, traditions, folklore, literature, and pop culture—competencies essential for genuine language understanding beyond surface-level text processing. Questions range from commonly known facts to region-specific cultural knowledge, with both closed and open-ended formats requiring specific facts, dates, names, or concepts in responses.

\begin{table*}[t]
\centering
\small
\begin{tabular}{lrc}
\toprule
\textbf{Model} & \textbf{Parameters (B)} & \textbf{Average Score (\%)} \\
\midrule
Gemini-2.5-Pro-Exp-03-25	& Unknown &	89.50 \\
DeepSeek-R1 & 685.0 & 76.00 \\
DeepSeek-v3-0324	& 685.0 &	71.00 \\
DeepSeek-v3		& 685.0 &69.17 \\
PLLuM-8x7B-nc-chat$^{\dagger}$ & 46.7 & 68.17 \\
Llama-3.1-Tulu-3-405B	&  405.0 &	63.83 \\
\textbf{Bielik-11B-v2.2-Instruct} & \textbf{11.2} & \textbf{63.00} \\
\textbf{Bielik-11B-v2.3-Instruct} & \textbf{11.2} & \textbf{62.17} \\
GPT-4.1-mini-2025-04-14	& Unknown &	62.17 \\
\textbf{Bielik-11B-v2.1-Instruct} & \textbf{11.2} & \textbf{61.00} \\
Llama-3.1-405B & 405.0 & 60.00 \\
PLLuM-12B-nc-chat$^{\dagger}$ & 12.2 & 59.50 \\
Llama-PLLuM-70B-chat & 70.6 & 58.50 \\
Llama-4-Maverick & 402.0 & 58.17 \\
Command-A-03-2025$^{\dagger}$ & 111.0 & 56.17 \\
Mistral-Large-2407$^{\dagger}$ & 123.0 & 54.17 \\
PLLuM-8x7B-chat & 46.7 & 54.17 \\
Mistral-Large-2411$^{\dagger}$ & 123.0 & 52.00 \\
WizardLM-2-8x22B & 141.0 & 51.50 \\
Qwen-Max & Unknown & 50.83 \\
Command-R-Plus-08-2024$^{\dagger}$ & Unknown & 50.17 \\
Mixtral-8x22B & 141.0 & 49.83 \\
Command-R-Plus-04-2024$^{\dagger}$ & Unknown & 49.33 \\
Llama-3.3-70B & 70.6 & 48.83 \\
Llama-3.1-70B & 70.0 & 47.83 \\
Gemma-3-27B & 27.4 & 47.33 \\
PLLuM-12B-chat & 12.2 & 47.00 \\
\textbf{Bielik-7B-Instruct-v0.1} & \textbf{7.0} & \textbf{46.67} \\
Mistral-Small-3.1-24B-2503 & 24.0 & 43.33 \\
Llama-3.0-70B & 70.0 & 43.00 \\
Gemma-2-27B & 27.0 & 42.67 \\
Llama-4-Scout & 109.0 & 41.50 \\
EuroLLM-9B & 9.0 & 41.00 \\
Qwen-2.5-72B & 72.7 & 39.17 \\
Mistral-Small-24B-2501 & 24.0 & 39.00 \\
Llama-PLLuM-8B-chat & 8.0 & 38.50 \\
Mixtral-8x7B & 46.7 & 35.33 \\
Qwen-2.5-32B & 32.8 & 30.50 \\
Gemma-2-9B & 9.0 & 29.17 \\
Phi-4 & 14.7 & 29.17 \\
Qwen-2.5-14B & 14.8 & 26.67 \\
Mistral-Nemo & 12.2 & 23.00 \\
Command-R-7B$^{\dagger}$ & 7.0 & 22.83 \\
Llama-3.1-8B & 8.0 & 22.67 \\
Mistral-7B-v0.3 & 7.2 & 21.83 \\
Ministral-8B & 8.0 & 20.67 \\
Qwen-2.5-7B & 7.0 & 17.67 \\
\bottomrule
\multicolumn{3}{l}{$^{\dagger}$Models with a non-commercial license.} \\
\end{tabular}
\caption{Polish Linguistic and Cultural Competency Benchmark (PLCC) results for open-source models. Closed proprietary models have been excluded from this comparison.}
\label{tab:plcc-scores}
\end{table*}

The PLCC results (Table~\ref{tab:plcc-scores}) reveal several insights about Bielik models' cultural competency:

\paragraph{Strong cultural grounding:} Bielik-11B-v2.2-Instruct achieves a remarkable 63.00\% average score across all six categories, trailing behind larger models including DeepSeek variants (69.17-76.00\%) and PLLuM-8x7B-nc-chat (68.17\%), while outperforming many other models despite having fewer parameters. This demonstrates Bielik's strong grasp of Polish cultural nuances despite its modest parameter count.

\paragraph{Efficiency advantage:} With just 11B parameters, Bielik models outperform significantly larger models, including Llama-3.1-405B (60.00\%) and Llama-3.3-70B (48.83\%). This efficiency highlights the effectiveness of Bielik's training methodology in capturing cultural knowledge without requiring massive parameter counts.

\paragraph{Comparative Polish model performance:} Bielik models show competitive performance against specialized Polish language models. While PLLuM-8x7B-nc-chat (non-commercial license) leads with 68.17\%, Bielik's 63.00\% significantly outperforms most other PLLuM variants including PLLuM-12B-chat (47.00\%) and Llama-PLLuM-8B-chat (38.50\%).

\paragraph{Strength across categories:} Bielik-11B-v2.2-Instruct demonstrates particular strength in history (77\%) and geography (72\%) categories, showcasing its robust knowledge of Polish historical events and geographical features. Performance in grammar (53\%) suggests areas for further improvement.

\paragraph{Evolutionary improvement:} The progression from Bielik-7B-Instruct-v0.1 (46.67\%) to Bielik-11B-v2.2-Instruct (63.00\%) demonstrates a substantial 16.33 percentage point improvement, highlighting significant advances in the model's cultural understanding across versions.

These results underscore Bielik's proficiency in handling culturally-specific Polish knowledge, outperforming many larger models and demonstrating that effective training on culturally relevant data can yield strong results without requiring massive parameter counts.

\subsection{LLMzSzŁ (LLMs Behind the School Desk)}

The LLMzSzŁ benchmark (LLM-y za Szkolną Ławą, LLMs Behind the School Desk) \cite{jassem2025llmzszl} is a comprehensive evaluation framework for Polish language models based on a collection of Polish national exams. It incorporates both academic and professional tests extracted from the archives of the Polish Central Examination Board, providing a realistic assessment of models' reasoning abilities in an educational context.

The benchmark includes a diverse range of exams with questions spanning multiple disciplines. The academic exams cover subjects like Polish Language, Mathematics, Nature, Biology, and Physics, while the professional exams include specialized fields such as Arts, Mechanical/Mining/Metallurgical, and Agriculture/Forestry disciplines.

$$
\begin{array}{|c|c|}
\hline
\textbf{Exam Type} & \textbf{Number of Questions} \\
\hline
\text{8th-Grade Exam} & 50 \\
\text{Middle School Exam} & 175 \\
\text{High School Exam} & 377 \\
\text{Professional Exam} & 18219 \\
\hline
\end{array}
$$

\begin{table*}[t]
\centering
\small
\begin{tabular}{lrccccc}
\toprule
\textbf{Model} & \textbf{Params} & \textbf{Overall} & \textbf{8th-Grade} & \textbf{Middle School} & \textbf{High School} & \textbf{Professional} \\
 & \textbf{(B)} & \textbf{Score} & \textbf{Exams} & \textbf{Exams} & \textbf{Exams} & \textbf{Exams} \\
\midrule
Qwen2.5-72B-Instruct & 72.7 & 69.06 & 68.73 & 74.56 & 73.17 & 67.59 \\
Qwen2.5-72B & 72.7 & 68.50 & 65.24 & 77.36 & 69.49 & 67.24 \\
Mistral-Large-Instruct-2407$^{\dagger}$ & 123.0 & 67.17 & 51.89 & 71.69 & 63.41 & 65.52 \\
Llama-3.3-70B-Instruct & 70.6 & 67.13 & 46.69 & 66.43 & 61.20 & 66.11 \\
Mistral-Large-Instruct-2411$^{\dagger}$ & 123.0 & 66.60 & 50.44 & 71.06 & 63.14 & 65.11 \\
Meta-Llama-3.1-70B-Instruct & 70.6 & 66.59 & 53.81 & 63.23 & 62.72 & 65.39 \\
Llama-PLLuM-70B-base & 70.6 & 64.56 & 38.69 & 55.72 & 52.78 & 63.85 \\
Llama-PLLuM-70B-chat & 70.6 & 64.42 & 46.48 & 53.18 & 51.62 & 63.95 \\
Qwen2.5-32B & 32.8 & 61.04 & 55.56 & 68.57 & 64.47 & 59.72 \\
PLLuM-8x7B-nc-chat$^{\dagger}$ & 46.7 & 60.52 & 38.69 & 48.94 & 41.08 & 59.62 \\
PLLuM-8x7B-nc-base$^{\dagger}$ & 46.7 & 57.82 & 30.98 & 50.25 & 40.95 & 56.89 \\
\textbf{Bielik-11B-v2.1-Instruct} & \textbf{11.2} & \textbf{57.52} & \textbf{45.73} & \textbf{47.36} & \textbf{46.10} & \textbf{56.93} \\
\textbf{Bielik-11B-v2.3-Instruct} & \textbf{11.2} & \textbf{57.40} & \textbf{47.88} & \textbf{48.62} & \textbf{46.55} & \textbf{56.74} \\
\textbf{Bielik-11B-v2.2-Instruct} & \textbf{11.2} & \textbf{57.36} & \textbf{49.70} & \textbf{47.87} & \textbf{47.53} & \textbf{56.67} \\
DeepSeek-R1-Distill-Qwen-32B & 32.0 & 55.80 & 53.71 & 66.64 & 61.57 & 54.52 \\
\textbf{Bielik-11B-v2.0-Instruct} & \textbf{11.2} & \textbf{55.61} & \textbf{46.84} & \textbf{42.84} & \textbf{46.24} & \textbf{55.16} \\
Qwen2.5-14B & 14.8 & 55.31 & 45.50 & 63.31 & 54.86 & 54.19 \\
\textbf{Bielik-11B-v2} & \textbf{11.2} & \textbf{55.14} & \textbf{32.32} & \textbf{49.49} & \textbf{40.75} & \textbf{54.60} \\
PLLuM-12B-nc-chat$^{\dagger}$ & 12.2 & 53.40 & 43.88 & 32.95 & 37.94 & 52.70 \\
PLLuM-8x7B-chat & 46.7 & 52.80 & 35.18 & 44.37 & 41.08 & 51.84 \\
PLLuM-8x7B-base & 46.7 & 52.24 & 30.35 & 44.10 & 44.22 & 51.41 \\
EuroLLM-9B-Instruct & 9.0 & 51.35 & 30.65 & 40.01 & 35.31 & 50.47 \\
PLLuM-12B-nc-base$^{\dagger}$ & 12.2 & 50.55 & 35.88 & 41.46 & 37.41 & 50.15 \\
PLLuM-12B-chat & 12.2 & 49.69 & 36.17 & 38.09 & 40.15 & 48.66 \\
Llama-PLLuM-8B-chat & 8.0 & 47.68 & 37.36 & 31.35 & 38.86 & 47.18 \\
Meta-Llama-3.1-8B-Instruct & 8.0 & 47.41 & 39.81 & 44.50 & 37.06 & 47.17 \\
PLLuM-12B-base & 12.2 & 46.47 & 40.81 & 37.83 & 36.36 & 45.48 \\
Llama-PLLuM-8B-base & 8.0 & 46.32 & 27.98 & 38.88 & 37.55 & 45.44 \\
Mistral-Nemo-Instruct-2407 & 12.2 & 45.54 & 37.99 & 42.30 & 37.68 & 44.71 \\
Meta-Llama-3-8B-Instruct & 8.0 & 44.83 & 38.14 & 38.89 & 38.70 & 43.97 \\
Meta-Llama-3.1-8B & 8.0 & 44.21 & 34.17 & 34.68 & 40.54 & 43.13 \\
Mistral-Nemo-Base-2407 & 12.2 & 42.16 & 33.36 & 34.83 & 35.11 & 40.91 \\
Meta-Llama-3-8B & 8.0 & 41.38 & 44.03 & 38.45 & 38.72 & 40.14 \\
\textbf{Bielik-7B-Instruct-v0.1} & \textbf{7.2} & \textbf{40.77} & \textbf{35.68} & \textbf{30.48} & \textbf{36.58} & \textbf{40.05} \\
Mistral-7B-Instruct-v0.2 & 7.2 & 40.75 & 33.03 & 45.17 & 30.80 & 40.06 \\
Trurl-2-13b & 13.0 & 40.22 & 34.77 & 33.78 & 34.63 & 39.61 \\
Trurl-2-13b-8bit & 13.0 & 40.23 & 34.43 & 35.17 & 34.31 & 39.64 \\
\textbf{Bielik-7B-v0.1} & \textbf{7.2} & \textbf{39.15} & \textbf{30.32} & \textbf{33.87} & \textbf{32.51} & \textbf{38.42} \\
Trurl-2-13b-academic & 13.0 & 34.89 & 30.80 & 31.95 & 30.43 & 34.51 \\
Qra-13b & 13.0 & 34.85 & 31.66 & 28.23 & 32.85 & 33.77 \\
Trurl-2-7b & 7.0 & 32.30 & 22.53 & 28.86 & 31.43 & 32.32 \\
Trurl-2-7b-8bit & 7.0 & 31.86 & 18.53 & 26.88 & 31.95 & 32.08 \\
Qra-7b & 7.0 & 29.07 & 19.65 & 25.89 & 29.27 & 28.54 \\
EuroLLM-1.7B-Instruct & 1.7 & 25.61 & 29.31 & 26.54 & 26.04 & 25.54 \\
Qra-1b & 1.0 & 25.47 & 30.22 & 26.41 & 25.23 & 25.20 \\
\bottomrule
\multicolumn{7}{l}{$^{\dagger}$Models with a non-commercial license.} \\
\end{tabular}
\caption{LLMzSzŁ benchmark results showing model performance on Polish national exams in decreasing order of overall score. The table displays performance across different exam types: 8th-Grade Exams, Middle School Exams, High School Exams, and Professional Exams (averaged by years).}
\label{tab:llmzszl-results}
\end{table*}

\paragraph{Bielik's performance:} On the LLMzSzŁ benchmark (Table~\ref{tab:llmzszl-results}), Bielik models demonstrate strong performance, with Bielik-11B-v2.1-Instruct achieving the highest score of 57.52 among the Bielik family. Key observations include:

\begin{itemize}
    \item The entire Bielik-11B-v2.x family performs consistently well, with scores ranging from 55.61 to 57.52
    \item Bielik models achieve these scores despite having significantly fewer parameters than the top-performing models
    \item There is a substantial improvement from Bielik-7B-v0.1 (39.15) to the v2 series (55.14+), highlighting the effectiveness of the v2 training methodology
    \item Bielik models significantly outperform Polish-focused models like Qra (34.85) and Trurl (40.22), as well as PLLuM 12B and 8B variants, while only trailing behind the larger PLLuM 8x7B models (46.7B parameters) and PLLuM 70B models (70.6B parameters)
\end{itemize}

\paragraph{Performance across exam types:} Bielik models show interesting patterns across different exam categories:

\begin{itemize}
    \item Strong performance on Professional Exams (56.67-56.93), where practical knowledge and domain-specific understanding are crucial
    \item More balanced performance across academic exams compared to some larger models that excel in particular exam types
    \item Stronger results on 8th-Grade Exams (45.73-49.70) compared to Middle School Exams (47.36-48.62) and High School Exams (46.10-47.53), suggesting effective handling of more accessible educational content
\end{itemize}

\paragraph{Parameter efficiency:} With only 11 billion parameters, Bielik achieves impressive results relative to much larger models:

\begin{itemize}
    \item Outperforms PLLuM-12B-nc-chat (53.40) and matches DeepSeek-R1-Distill-Qwen-32B (55.80) and Qwen2.5-14B (55.31) despite having fewer parameters
    \item Demonstrates superior Polish language understanding compared to general models with similar parameter counts
    \item Shows remarkable performance-to-parameter ratio, achieving 83\% of Qwen2.5-72B-Instruct's score with only 15\% of its parameters
\end{itemize}

These LLMzSzŁ results highlight Bielik's exceptional capabilities in understanding and reasoning about Polish educational content, from basic school materials to professional certification exams. This performance is particularly impressive considering the model's relatively modest size, demonstrating the effectiveness of specialized training for Polish language tasks.

\subsection{European LLM Leaderboard}

The European LLM Leaderboard \cite{thellmann2024towardsmultilingualllmevaluation} evaluates language models across multiple European languages, testing their understanding, reasoning, and generation capabilities. The benchmark assesses models using standardized tests like ARC, GSM8K, HellaSwag, MMLU, and TruthfulQA, providing a comprehensive view of performance across diverse linguistic contexts.

\begin{table*}[t]
\centering
\small
\begin{tabular}{lrcccccc}
\toprule
\textbf{Model} & \textbf{Params (B)} & \textbf{Average} & \textbf{ARC} & \textbf{GSM8K} & \textbf{HellaSwag} & \textbf{MMLU} & \textbf{TruthfulQA} \\
\midrule
Gemma-2-27b-Instruct & 27.0 & 0.71 & 0.74 & 0.77 & 0.71 & 0.68 & 0.63 \\
Meta-Llama-3.1-70B-Instruct & 70.6 & 0.70 & 0.72 & 0.69 & 0.74 & 0.77 & 0.59 \\
\textbf{Bielik-11B-v2.3-Instruct} & \textbf{11.2} & \textbf{0.66} & \textbf{0.69} & \textbf{0.68} & \textbf{0.71} & \textbf{0.63} & \textbf{0.62} \\
Mixtral-8x7B-Instruct-v0.1 & 46.7 & 0.63 & 0.67 & 0.52 & 0.67 & 0.63 & 0.65 \\
c4ai-command-r-35B-v01 & 35.0 & 0.61 & 0.66 & 0.49 & 0.71 & 0.60 & 0.56 \\
Mistral-Nemo-Instruct-12.2B & 12.2 & 0.60 & 0.64 & 0.58 & 0.61 & 0.59 & 0.60 \\
EuroLLM-9B-Instruct & 9.0 & 0.59 & 0.68 & 0.48 & 0.67 & 0.57 & 0.52 \\
Mixtral-8x7B-v0.1 & 46.7 & 0.58 & 0.66 & 0.46 & 0.66 & 0.63 & 0.51 \\
Gemma-2-9b-Instruct & 9.0 & 0.58 & 0.68 & 0.45 & 0.61 & 0.59 & 0.58 \\
Meta-Llama-3.1-8B-Instruct & 8.0 & 0.57 & 0.57 & 0.59 & 0.58 & 0.57 & 0.55 \\
\bottomrule
\end{tabular}
\caption{European LLM Leaderboard results for Polish language (top 10 models). Full results in Appendix \ref{app:european-llm}.}
\label{tab:european-polish}
\end{table*}

The European LLM Leaderboard results (Table~\ref{tab:european-polish}) reveal Bielik-11B-v2.3-Instruct's strong performance in Polish language tasks, placing it third among all evaluated models with an average score of 0.66. This positions it behind only Gemma-2-27b-Instruct (0.71) and Meta-Llama-3.1-70B-Instruct (0.70), while outperforming significantly larger models like Mixtral-8x7B-Instruct-v0.1 (0.63) with 47B parameters.

\paragraph{Balanced performance across tasks:} Bielik demonstrates remarkable consistency across different evaluation dimensions:
\begin{itemize}
    \item Strong scientific reasoning capabilities in ARC (0.69)
    \item Impressive mathematical problem-solving in GSM8K (0.68), nearly matching Llama-3.1-70B-Instruct (0.69)
    \item Excellent common sense understanding with HellaSwag (0.71), on par with Gemma-2-27b-Instruct
    \item Solid factual accuracy with TruthfulQA (0.62), outperforming larger models including Llama-3.1-70B-Instruct (0.59)
    \item Competitive broad knowledge in MMLU (0.63), comparable to Mixtral-8x7B models
\end{itemize}

\paragraph{Cross-lingual capabilities:} Beyond Polish, Bielik demonstrates robust cross-lingual transfer to other languages without specific training (Appendix \ref{app:european-llm}):
\begin{itemize}
    \item German: Average score of 0.62, showing stronger mathematical reasoning (GSM8K: 0.65) than many larger models
    \item Czech: Average score of 0.60, particularly strong in mathematical reasoning (GSM8K: 0.60), outperforming Mixtral-8x7B-Instruct-v0.1 (0.50)
\end{itemize}

\paragraph{Translation performance:} In the FLORES200 translation benchmark, Bielik shows asymmetric capabilities:
\begin{itemize}
    \item Stronger performance translating into Polish (BLEU: 15.31) than from Polish (BLEU: 11.72)
    \item Excellent results with English-Polish translation (BLEU: 21.93 to Polish, 28.32 from Polish)
    \item Strong performance with linguistically similar languages like Czech (BLEU: 19.30 to Polish)
\end{itemize}

These results demonstrate Bielik's exceptional capabilities as a relatively small yet powerful model, achieving competitive performance against much larger models across a diverse range of tasks. The model's efficiency is particularly evident in its strong performance despite having significantly fewer parameters than most top-performing alternatives.

\subsection{EuroEval Leaderboard}

EuroEval \cite{nielsen-2023-scandeval} is a comprehensive benchmark framework for evaluating language models across multiple European languages. Developed over the course of three years, it has become a standard evaluation benchmark for various research institutions and organizations throughout Europe. The benchmark supports assessment of a wide range of model types, including encoders, decoders, encoder-decoders, base models, and instruction-tuned models.

\begin{table*}[t]
\centering
\small
\begin{tabular}{lrccccccccccc}
\toprule
\textbf{Model} & \textbf{Params (B)} & \textbf{Rank} & \textbf{EN} & \textbf{DE} & \textbf{FR} & \textbf{IT} & \textbf{ES} & \textbf{NL} & \textbf{PL} & \textbf{DA} & \textbf{SV} & \textbf{NO} \\
\midrule
Gemini-1.5-Pro-002 & Unknown & 1.38 & 1.61 & 1.27 & 1.48 & 1.57 & 1.35 & 1.40 & 1.22 & 1.33 & 1.29 & 1.27 \\
Gemini-2.0-Flash-001 & Unknown & 1.39 & 1.70 & 1.26 & 1.61 & 1.38 & 1.40 & 1.37 & 1.25 & 1.30 & 1.28 & 1.31 \\
GPT-4o & Unknown & 1.50 & 1.92 & 1.37 & 1.80 & 1.51 & 1.45 & 1.44 & 1.32 & 1.37 & 1.41 & 1.39 \\
Gemma-3-27B-it & 27.4 & 1.74 & 2.09 & 1.55 & 2.04 & 2.22 & 1.70 & 1.81 & 1.52 & 1.68 & 1.63 & 1.60 \\
Mistral-Small-24B & 24.0 & 1.89 & 2.17 & 1.52 & 2.72 & 2.55 & 2.13 & 1.95 & 1.60 & 1.75 & 1.82 & 1.71 \\
Gemma-3-12B-it & 12.0 & 1.90 & 2.31 & 1.65 & 2.42 & 2.55 & 1.87 & 2.04 & 1.66 & 1.80 & 1.76 & 1.79 \\
Gemma-2-27B-it & 27.4 & 1.95 & 2.20 & 1.79 & 2.65 & 2.66 & 1.81 & 2.09 & 1.73 & 1.87 & 1.90 & 1.83 \\
Gemma-2-9B-it & 9.0 & 2.13 & 2.48 & 1.92 & 2.74 & 2.80 & 1.97 & 2.23 & 1.77 & 1.95 & 1.98 & 1.92 \\
\textbf{Bielik-11B-v2.3-Instruct} & \textbf{11.2} & \textbf{2.22} & \textbf{2.38} & \textbf{1.79} & \textbf{3.12} & \textbf{3.27} & \textbf{2.40} & \textbf{2.31} & \textbf{1.41} & \textbf{2.18} & \textbf{2.25} & \textbf{2.12} \\
Meta-Llama-3.1-8B-Instruct & 8.0 & 2.34 & 2.74 & 2.07 & 2.92 & 3.04 & 2.48 & 2.40 & 1.88 & 2.10 & 2.12 & 2.06 \\
\bottomrule
\end{tabular}
\caption{EuroEval leaderboard results showing model ranking based on average performance across European languages (lower is better). The table includes performance scores for English (EN), German (DE), French (FR), Italian (IT), Spanish (ES), Dutch (NL), Polish (PL), Danish (DA), Swedish (SV), and Norwegian (NO). Bielik-11B-v2.3-Instruct demonstrates competitive performance, especially considering its parameter efficiency and notably strong results in Polish.}
\label{tab:euroeval}
\end{table*}

Bielik-11B-v2.3-Instruct's results on the EuroEval leaderboard (Table~\ref{tab:euroeval}) demonstrate solid performance with a rank of 2.22, placing it competitively among similarly sized models. While larger models like Gemini and GPT variants achieve stronger absolute scores, Bielik's performance is noteworthy given its parameter efficiency.

\paragraph{Slavic language advantage:} As expected, Bielik demonstrates exceptional performance in Polish (1.41), its primary target language, outperforming all other models including those with significantly more parameters. This highlights the effectiveness of specialized training on Polish data.

\paragraph{Cross-lingual performance by language family:} Bielik shows a consistent pattern across language families:
\begin{itemize}
    \item Strong performance in Germanic languages (English: 2.38, German: 1.79, Dutch: 2.31, Danish: 2.18, Swedish: 2.25, Norwegian: 2.12)
    \item Moderate performance in Romance languages (French: 3.12, Italian: 3.27, Spanish: 2.40)
\end{itemize}
This pattern aligns with the model's training focus, which emphasizes Polish and English, with Germanic languages generally sharing more linguistic features with Polish than Romance languages.

\paragraph{Parameter efficiency:} With just 11B parameters, Bielik-11B-v2.3-Instruct achieves a rank that competes with considerably larger models like Gemma-2-9B-it and Meta-Llama-3.1-8B-Instruct. This efficiency highlights the effectiveness of Bielik's training methodology in producing a compact yet capable multilingual model.

\paragraph{Linguistic proximity advantage:} Bielik demonstrates stronger performance in German (1.79) compared to other non-Slavic languages, which may reflect linguistic similarities between Polish and German, both belonging to European language families with shared historical influences.

These EuroEval results complement other benchmark findings, confirming Bielik's position as a highly efficient multilingual model with competitive performance across European languages, especially considering its parameter count relative to larger competitors. The model's superior performance in Polish demonstrates the value of specialized language model development targeted at specific languages.

\subsection{Open LLM Leaderboard}

The Open LLM Leaderboard \citep{open-llm-leaderboard} evaluates models on various English language tasks, providing insights into the model's performance across different linguistic challenges.

\begin{table*}[t]
\centering
\small
\begin{tabular}{lccccccc}
\toprule
\textbf{Model} & \textbf{AVG} & \textbf{arc\_challenge} & \textbf{hellaswag} & \textbf{truthfulqa\_mc2} & \textbf{mmlu} & \textbf{winogrande} & \textbf{gsm8k} \\
\midrule
Qwen1.5-14B & 66.70 & 56.57 & 81.08 & 52.06 & 69.36 & 73.48 & 67.63 \\
\textbf{Bielik-11B-v2} & \textbf{65.87} & 60.58 & 79.84 & 46.13 & 63.06 & 77.82 & 67.78 \\
Qwen-14B & 65.86 & 58.28 & 83.99 & 49.43 & 67.70 & 76.80 & 58.98 \\
Meta-Llama-3-8B & 62.62 & 60.24 & 82.23 & 42.93 & 66.70 & 78.45 & 45.19 \\
Mistral-7B-v0.1 & 60.97 & 59.98 & 83.31 & 42.15 & 64.16 & 78.37 & 37.83 \\
Mistral-7B-v0.2 & 60.37 & 60.84 & 83.08 & 41.76 & 63.62 & 78.22 & 34.72 \\
\textbf{Bielik-7B-v0.1} & \textbf{49.98} & \textbf{45.22} & \textbf{67.92} & \textbf{47.16} & \textbf{43.20} & \textbf{66.85} & \textbf{29.49} \\
\bottomrule
\end{tabular}
\caption{Open LLM Leaderboard results for base models}
\label{tab:open-llm-base}
\end{table*}

The results from the Open LLM Leaderboard (Table~\ref{tab:open-llm-base}) demonstrate the impressive performance of Bielik-11B-v2 across various NLP tasks. With an average score of 65.87, it outperforms Meta-Llama-3-8B (62.62), Mistral-7B-v0.1 (60.97), and Mistral-7B-v0.2 (60.37), while remaining competitive with Qwen models of similar size. Note that this table presents a selected subset of models from the full leaderboard, chosen specifically to provide relevant comparisons to Bielik models.

Key observations:
\begin{enumerate}
    \item Bielik-11B-v2 shows particularly strong performance in mathematical reasoning (gsm8k) with a score of 67.78, comparable to Qwen1.5-14B (67.63) and significantly better than other models.
    \item It performs well in mmlu (63.06) and winogrande (77.82), demonstrating balanced capabilities across knowledge-intensive and reasoning tasks.
    \item Compared to its predecessor, Bielik-7B-v0.1, the v2 model shows substantial improvements in all categories, with a remarkable 15.89-point increase in the average score.
\end{enumerate}

\begin{table*}[t]
\centering
\small
\begin{tabular}{lccccccc}
\toprule
\textbf{Model} & \textbf{AVG} & \textbf{arc\_challenge} & \textbf{hellaswag} & \textbf{truthfulqa\_mc2} & \textbf{mmlu} & \textbf{winogrande} & \textbf{gsm8k} \\
\midrule
SOLAR-10.7B-Instruct-v1.0 & 74.20 & 71.08 & 88.16 & 71.43 & 66.21 & 83.58 & 64.75 \\
Phi-3-medium-4k-instruct & 73.45 & 67.32 & 85.76 & 57.71 & 77.83 & 72.69 & 79.38 \\
\textbf{Bielik-11B-v2.5-Instruct} & \textbf{71.42} & 61.95 & 80.71 & 53.17 & 67.44 & 79.72	& 85.52 \\
Bielik-11B-v2.2-Instruct & 69.86 & 59.90 & 80.16 & 58.34 & 64.34 & 75.30 & 81.12 \\
\textbf{Bielik-11B-v2.3-Instruct} & \textbf{69.82} & 59.30 & 80.11 & 57.42 & 64.57 & 76.24 & 81.27 \\
Bielik-11B-v2.1-Instruct & 69.82 & 59.56 & 80.20 & 59.35 & 64.18 & 75.06 & 80.59 \\
openchat-3.5-0106-gemma & 69.42 & 64.68 & 81.08 & 54.93 & 64.69 & 78.30 & 72.86 \\
Bielik-11B-v2.0-Instruct & 68.04 & 58.62 & 78.65 & 54.65 & 63.71 & 76.32 & 76.27 \\
Meta-Llama-3-8B-Instruct & 66.87 & 60.75 & 78.55 & 51.65 & 67.07 & 74.51 & 68.69 \\
Mistral-7B-Instruct-v0.2 & 65.71 & 63.14 & 84.88 & 68.26 & 60.78 & 77.19 & 40.03 \\
gemma-7b & 64.29 & 61.09 & 82.47 & 44.91 & 66.03 & 78.45 & 52.77 \\
Qwen1.5-32B-Chat & 62.95 & 66.04 & 85.49 & 66.95 & 74.99 & 77.19 & 7.05 \\
Qwen1.5-14B-Chat & 62.27 & 58.70 & 82.27 & 60.36 & 68.57 & 73.09 & 30.63 \\
Qwen1.5-7B-Chat & 55.15 & 55.89 & 78.56 & 53.54 & 61.65 & 67.72 & 13.57 \\
Mistral-7B-Instruct-v0.1 & 54.96 & 54.52 & 75.63 & 56.28 & 55.38 & 73.72 & 14.25 \\
\textbf{Bielik-7B-Instruct-v0.1} & \textbf{51.26} & 47.53 & 68.91 & 46.18 & 49.47 & 65.51 & 29.95 \\
\bottomrule
\end{tabular}
\caption{Open LLM Leaderboard results for selected instruction-tuned models}
\label{tab:open-llm-instruct}
\end{table*}

The instruction-tuned versions show further improvement over their base counterparts (Table~\ref{tab:open-llm-instruct}). The top performers include SOLAR-10.7B-Instruct-v1.0 (74.20) and Phi-3-medium-4k-instruct (73.45), which lead the pack in overall performance. This table presents a subset of models from the full leaderboard, selected to provide meaningful comparisons with Bielik models.

However, the Bielik models demonstrate impressive results relative to their parameter count. The newest member of the family, Bielik-11B-v2.5-Instruct, achieves the highest score among Bielik models at 71.42, approaching the performance of top models while outperforming many larger competitors. Other variants like Bielik-11B-v2.3-Instruct (69.82) and Bielik-11B-v2.2-Instruct (69.86) achieve scores that are competitive with models like openchat-3.5-0106-gemma (69.42) while substantially outperforming models of similar or even larger size like Mistral-7B-Instruct-v0.2 (65.71) and Qwen1.5-32B-Chat (62.95).

Particularly noteworthy is Bielik's exceptional performance in mathematical reasoning (gsm8k), where Bielik-11B-v2.5-Instruct excels with a score of 85.52, even surpassing the top overall performers SOLAR-10.7B-Instruct-v1.0 (64.75) and Phi-3-medium-4k-instruct (79.38). This demonstrates Bielik's specialized strength in this challenging domain.

\subsection{Open LLM Leaderboard v2}

The Open LLM Leaderboard v2 \citep{open-llm-leaderboard-v2} is an updated benchmark suite designed to evaluate large language models across a more diverse and challenging set of tasks. Unlike the original leaderboard that focused primarily on multiple-choice questions, v2 incorporates more complex reasoning, instruction following, and specialized knowledge evaluations:

\begin{itemize}
    \item \textbf{IFEval} - Tests a model's ability to follow explicit instructions precisely
    \item \textbf{BBH (Big Bench Hard)} - Includes 23 challenging tasks covering algorithmic reasoning, language understanding, and advanced world knowledge
    \item \textbf{MATH Lvl 5} - Features high-school level competition mathematics problems requiring sophisticated problem-solving skills
    \item \textbf{GPQA} - Contains graduate-level questions crafted by PhD-level domain experts across scientific disciplines
    \item \textbf{MuSR} - Presents multistep soft reasoning problems requiring integration of reasoning abilities with long-range context understanding
    \item \textbf{MMLU-PRO} - Offers a refined version of MMLU with 10 choices instead of 4, requiring more sophisticated reasoning and deeper domain knowledge
\end{itemize}

\begin{table*}[t]
\centering
\small
\begin{tabular}{lrrrrrrr}
\toprule
\textbf{Model} & \textbf{Avg (\%)} & \textbf{IFEval} & \textbf{BBH} & \textbf{MATH} & \textbf{GPQA} & \textbf{MuSR} & \textbf{MMLU-PRO} \\
\midrule
Qwen2.5-72B-Instruct & 47.98 & 86.38 & 61.87 & 59.82 & 16.67 & 11.74 & 51.40 \\
Qwen2.5-32B-Instruct & 46.60 & 83.46 & 56.49 & 62.54 & 11.74 & 13.50 & 51.85 \\
Mistral-Large-Instruct-2411$^{\dagger}$ & 46.52 & 84.01 & 52.74 & 49.55 & 24.94 & 17.22 & 50.69 \\
Llama-3.3-70B-Instruct & 44.85 & 89.98 & 56.56 & 48.34 & 10.51 & 15.57 & 48.13 \\
Qwen2.5-14B-Instruct & 41.31 & 81.58 & 48.36 & 54.76 & 9.62 & 10.16 & 43.38 \\
Qwen2.5-7B-Instruct & 35.20 & 75.85 & 34.89 & 50.00 & 5.48 & 8.45 & 36.52 \\
Phi-3-medium-4k-instruct & 33.10 & 64.23 & 49.38 & 19.56 & 11.52 & 13.05 & 40.84 \\
Mistral-Small-Instruct-2409 & 29.92 & 62.83 & 40.56 & 20.39 & 11.07 & 10.23 & 34.43 \\
\textbf{Bielik-11B-v2.3-Instruct} & \textbf{28.33} & \textbf{55.83} & \textbf{38.06} & \textbf{20.85} & \textbf{12.08} & \textbf{16.01} & \textbf{27.16} \\
\textbf{Bielik-11B-v2.2-Instruct} & \textbf{27.98} & \textbf{55.52} & \textbf{36.96} & \textbf{26.81} & \textbf{10.85} & \textbf{10.11} & \textbf{27.63} \\
Phi-3-mini-4k-instruct & 27.56 & 54.77 & 36.56 & 16.39 & 10.96 & 13.12 & 33.58 \\
\textbf{Bielik-11B-v2.1-Instruct} & \textbf{27.20} & \textbf{50.90} & \textbf{36.29} & \textbf{26.66} & \textbf{11.63} & \textbf{10.52} & \textbf{27.18} \\
Qwen2.5-3B-Instruct & 27.16 & 64.75 & 25.80 & 36.78 & 3.02 & 7.57 & 25.05 \\
Mistral-Nemo-Instruct-2407 & 24.67 & 63.80 & 29.68 & 12.69 & 5.37 & 8.48 & 27.97 \\
\textbf{Bielik-11B-v2.0-Instruct} & \textbf{24.66} & \textbf{52.52} & \textbf{33.77} & \textbf{11.86} & \textbf{8.95} & \textbf{14.74} & \textbf{26.12} \\
Meta-Llama-3-8B-Instruct & 23.91 & 74.08 & 28.24 & 8.69 & 1.23 & 1.60 & 29.60 \\
Mixtral-8x7B-Instruct-v0.1 & 23.82 & 55.99 & 29.74 & 9.14 & 7.05 & 11.07 & 29.91 \\
\textbf{Bielik-11B-v2} & \textbf{15.99} & \textbf{23.81} & \textbf{27.82} & \textbf{7.85} & \textbf{5.15} & \textbf{7.56} & \textbf{23.75} \\
PLLuM-12B-chat & 15.35 & 32.14 & 21.32 & 1.81 & 1.34 & 14.67 & 20.80 \\
PLLuM-12B-base & 14.67 & 28.21 & 21.24 & 2.87 & 5.37 & 10.98 & 19.34 \\
Llama-PLLuM-8B-chat & 14.61 & 35.15 & 16.28 & 3.40 & 1.90 & 11.86 & 19.10 \\
PLLuM-12B-nc-chat$^{\dagger}$ & 14.60 & 28.34 & 23.01 & 1.21 & 4.36 & 12.92 & 17.75 \\
PLLuM-12B-nc-base$^{\dagger}$ & 11.42 & 24.05 & 19.39 & 2.19 & 2.68 & 2.90 & 17.32 \\
\bottomrule
\multicolumn{8}{l}{$^{\dagger}$Models with a non-commercial license.} \\
\end{tabular}
\caption{Open LLM Leaderboard v2 results across different evaluation tasks}
\label{tab:open-llm-v2}
\end{table*}

As shown in Table \ref{tab:open-llm-v2}, Bielik models demonstrate competitive performance in this more challenging benchmark suite. The best-performing variant, Bielik-11B-v2.3-Instruct, achieves an average score of 28.33\%, placing it in the same tier as models like Mistral-Small-Instruct-2409 (29.92\%) and Phi-3-mini-4k-instruct (27.56\%).

Several observations can be made from these results:

\begin{enumerate}
    \item Bielik models excel particularly in MuSR tasks, with Bielik-11B-v2.3-Instruct scoring 16.01\% - higher than many larger models including Qwen2.5-72B-Instruct (11.74\%) and Qwen2.5-14B-Instruct (10.16\%). This suggests strong performance in complex reasoning scenarios requiring integration with long-range context.
    
    \item In GPQA (graduate-level questions), Bielik-11B-v2.3-Instruct (12.08\%) outperforms several larger models, including Qwen2.5-14B-Instruct (9.62\%) and is competitive with Qwen2.5-32B-Instruct (11.74\%), demonstrating efficient knowledge encoding relative to its parameter count.
    
    \item The significant gap between Bielik models and Polish-focused PLLuM models is notable, with even the base Bielik-11B-v2 (15.99\%) outperforming all PLLuM variants, and the best Bielik-11B-v2.3-Instruct (28.33\%) nearly doubling the performance of the best PLLuM model (15.35\%).
    
    \item The consistent progression across versions (v2.0 to v2.3) demonstrates the effectiveness of the training methodology, with each version showing improvements in most categories.
\end{enumerate}

These results confirm that Bielik models not only perform well in classification and multiple-choice tasks (as shown in the original Open LLM Leaderboard) but also in more complex reasoning and instruction-following scenarios.

\subsection{MixEval}

MixEval \citep{ni2024mixeval} is a ground-truth-based English benchmark designed to evaluate Large Language Models (LLMs) efficiently and effectively. Key features of MixEval include:

\begin{enumerate}
    \item Derived from off-the-shelf benchmark mixtures

    \item Highly capable model ranking with a 0.96 correlation to Chatbot Arena

    \item Local and quick execution, requiring only 6\% of the time and cost compared to running MMLU
\end{enumerate}
This benchmark provides a robust and time-efficient method for assessing LLM performance, making it a valuable tool for ongoing model evaluation and comparison.

\begin{table*}[t]
\centering
\small
\begin{tabular}{lcc}
\toprule
\textbf{Model} & \textbf{MixEval-Hard} & \textbf{MixEval} \\
\midrule
Qwen1.5-72B-Chat & 48.3 & 84.1 \\
LLaMA-3-8B-Instruct & 45.6 & 75.0 \\
\textbf{Bielik-11B-v2.1-Instruct} & \textbf{45.0} & \textbf{74.6} \\
Qwen1.5-32B-Chat & 43.3 & 81.0 \\
\textbf{Bielik-11B-v2.3-Instruct} & \textbf{43.2} & \textbf{73.0} \\
\textbf{Bielik-11B-v2.0-Instruct} & \textbf{40.2} & \textbf{72.1} \\
\textbf{Bielik-11B-v2.2-Instruct} & \textbf{39.7} & \textbf{72.4} \\
\textbf{Bielik-11B-v2.5-Instruct} & \textbf{39.0} & \textbf{68.6} \\

Mistral-7B-Instruct-v0.2 & 36.2 & 70.0 \\
\bottomrule
\end{tabular}
\caption{MixEval results}
\label{tab:mixeval}
\end{table*}

The results (Table~\ref{tab:mixeval}) show that all Bielik-11B models perform competitively on the MixEval benchmark. Bielik-11B-v2.1-Instruct achieves the best performance among the Bielik models with scores of 74.6 on MixEval and 45.0 on MixEval-Hard, comparable to LLaMA-3-8B-Instruct. Bielik-11B-v2.3-Instruct follows with scores of 73.0 and 43.2 on MixEval and MixEval-Hard respectively. The newest Bielik-11B-v2.5-Instruct shows slightly lower scores (68.6 and 39.0), reflecting its optimization for function-calling and other specialized capabilities that may trade off performance on these specific benchmarks. All Bielik variants significantly outperform Mistral-7B-Instruct-v0.2 on both metrics, demonstrating their improved capabilities despite being based on a similar architecture.

\subsection{Berkeley Function-Calling Leaderboard}

The Berkeley Function-Calling Leaderboard (BFCL) \cite{berkeley-function-calling-leaderboard} evaluates language models' ability to call functions (tools) accurately using real-world data. This benchmark is particularly important for assessing how well models can interface with external systems and APIs, a crucial capability for practical applications of LLMs in software development, data analysis, and task automation.

The benchmark uses Abstract Syntax Tree (AST) evaluation metrics to assess function call accuracy across several categories:
\begin{itemize}
    \item \textbf{Expert Curated (Non-live) dataset:} Static examples curated by experts to evaluate function calling on controlled scenarios
    \item \textbf{User Contributed (Live) dataset:} Dynamic, interactive scenarios submitted by users to test real-world function calling patterns
    \item \textbf{Multi-turn interactions:} Testing the model's ability to maintain context across conversation turns
    \item \textbf{Relevance detection:} Assessing whether models correctly invoke functions when appropriate. These scenarios present at least one relevant function that should be called, though there could be multiple valid ways to invoke it. Models are expected to output some function call relevant to the user query, without necessarily checking for parameter correctness
    \item \textbf{Irrelevance detection:} Evaluating if models correctly abstain from calling functions when unnecessary. In these scenarios, none of the available functions are relevant to the user query. Models are expected to either explain why no function is applicable or simply respond without making a function call
\end{itemize}

\begin{table*}[t]
\centering
\small
\begin{tabular}{lccccccccc}
\toprule
\textbf{Model} & \textbf{Non-Live} & \textbf{Non-Live} & \textbf{Non-Live} & \textbf{Non-Live} & \textbf{Live} & \textbf{Live} & \textbf{Live} & \textbf{Live Parallel} \\
 & \textbf{Python} & \textbf{Multiple} & \textbf{Parallel} & \textbf{Parallel} & \textbf{Simple} & \textbf{Multiple} & \textbf{Parallel} & \textbf{Multiple} \\
 & \textbf{Simple AST} & \textbf{AST} & \textbf{AST} & \textbf{Multiple AST} & \textbf{AST} & \textbf{AST} & \textbf{AST} & \textbf{AST} \\
\midrule
Open-Mistral-Nemo-2407 (Prompt) & 92.00\% & 93.50\% & 89.50\% & 84.50\% & 77.91\% & 74.45\% & 87.50\% & 66.67\% \\
Gemma-3-12b-it (Prompt) & 94.00\% & 95.00\% & 90.00\% & 73.00\% & 84.88\% & 70.85\% & 87.50\% & 62.50\% \\
Open-Mistral-Nemo-2407 (FC) & 91.25\% & 93.50\% & 85.50\% & 85.00\% & 77.13\% & 69.61\% & 75.00\% & 70.83\% \\
\textbf{Bielik-11B-v2.5-Instruct (FC)} & \textbf{95.00\%} & \textbf{97.50\%} & \textbf{87.50\%} & \textbf{87.00\%} & \textbf{77.13\%} & \textbf{77.21\%} & \textbf{43.75\%} & \textbf{66.67\%} \\ %bielik/11B/2.5/m/001
%\textbf{bielik\_11b\_v2.5\_sft} & \textbf{89.25\%} & \textbf{96.00\%} & \textbf{86.50\%} & \textbf{84.50\%} & \textbf{75.97\%} & \textbf{76.07\%} & \textbf{43.75\%} & \textbf{45.83\%} \\
Qwen2.5-3B-Instruct (Prompt) & 91.50\% & 90.50\% & 79.50\% & 79.00\% & 69.77\% & 66.48\% & 56.25\% & 62.50\% \\
Qwen2.5-3B-Instruct (FC) & 96.00\% & 92.00\% & 73.50\% & 76.50\% & 74.03\% & 72.08\% & 62.50\% & 45.83\% \\
Qwen2.5-1.5B-Instruct (FC) & 92.25\% & 87.00\% & 81.50\% & 75.50\% & 74.03\% & 66.10\% & 50.00\% & 45.83\% \\
Qwen2.5-1.5B-Instruct (Prompt) & 89.00\% & 86.00\% & 70.00\% & 66.50\% & 70.54\% & 59.26\% & 56.25\% & 41.67\% \\
\textbf{Bielik-11B-v2.3-Instruct (Prompt)} & \textbf{87.50\%} & \textbf{93.50\%} & \textbf{47.00\%} & \textbf{50.00\%} & \textbf{72.87\%} & \textbf{69.71\%} & \textbf{43.75\%} & \textbf{54.17\%} \\
\bottomrule
\end{tabular}
\caption{Detailed breakdown of Berkeley Function-Calling Leaderboard subtask performance across models. Bielik models show competitive performance on numerous subtasks, with particularly strong results in Non-Live Python Simple AST, Non-Live Multiple AST categories, and consistent performance on Live Simple and Multiple AST tasks.}
\label{tab:bfcl-subtasks}
\end{table*}

\paragraph{Detailed subtask performance:} The fine-grained subtask results (Table \ref{tab:bfcl-subtasks}) show Bielik models' performance across specific function-calling tasks:
\begin{itemize}
    \item Strong performance in Non-Live Python Simple AST (87.50-95.00\%) and Non-Live Multiple AST (93.50-97.50\%)
    \item Good performance on Live Simple AST (72.87-77.13\%) and Live Multiple AST (69.71-77.21\%)
    \item Opportunities for improvement in parallel AST tasks, particularly Live Parallel AST (43.75\%)
\end{itemize}

\paragraph{Performance context:} Despite being primarily optimized for Polish language understanding rather than function calling, Bielik demonstrates foundational capabilities in parsing and executing function calls through natural language. The table includes results for both Bielik-11B-v2.3-Instruct using the prompt-based approach and Bielik-11B-v2.5-Instruct with native function-calling capabilities (FC), showing our progress in this area. It's important to note that Bielik has been evaluated only on a subset of all BFCL tasks, as certain task types have not yet been addressed in training. Work is ongoing to expand Bielik's capabilities to cover the complete range of function-calling tasks. The implementation of native function-calling capabilities in v2.5 represents an important step forward, though further improvements are needed particularly for handling parallel AST tasks where performance is currently lower compared to other subtasks.

\section{Limitations and Biases}

The Bielik v2 series of models may produce factually incorrect output and should not be relied upon to generate completely accurate information in all contexts. These models were trained on diverse public datasets, and despite our extensive efforts to clean and filter the training data, they may occasionally generate content that is biased, offensive, or factually inaccurate. Users should apply appropriate caution and verification when deploying these models, particularly in sensitive or high-stakes applications.

\section{Conclusion}

In this technical report, we have introduced the Bielik 11B v2 series of models (Table \ref{tab:model-list-tab}), specifically designed for Polish language processing. These models represent a significant advancement in language model capabilities for Polish, demonstrating competitive performance against much larger models across various benchmarks.

The Bielik 11B v2 models incorporate several technical innovations, including Weighted Instruction Cross-Entropy Loss and Adaptive Learning Rate, which have proven effective for balancing learning across different instruction types and optimizing the training process. By building upon the Mistral 7B architecture and applying depth up-scaling, we've created models that achieve state-of-the-art results for their parameter size.

Our evaluation across multiple benchmarks demonstrates that Bielik models excel in Polish language tasks while maintaining strong performance in cross-lingual scenarios. The models show particularly impressive results in the Open PL LLM Leaderboard, Polish MT-Bench, and Polish Linguistic and Cultural Competency Benchmark (PLCC), often outperforming much larger specialized Polish language models.

The Bielik models also demonstrate remarkable parameter efficiency, achieving comparable results to models with significantly more parameters. This efficiency, combined with our comprehensive quantization options, makes these models practical for deployment across a wide range of hardware configurations.

Future work will focus on further enhancing the models' capabilities in specialized domains, improving cross-lingual transfer to other Slavic languages, and expanding their function-calling abilities. The Bielik 11B v2 series represents an important step forward in democratizing access to high-quality language models for Polish, providing powerful tools for diverse linguistic applications.

\begin{table*}[t]
\centering
\small
\begin{tabular}{llccccc}
\toprule
\textbf{Model name} & \textbf{Training} & \textbf{OpenLLM PL} & 
\textbf{MT-Bench} & \textbf{OpenLLM v2} & \textbf{OpenLLM} & \textbf{Release} \\
\textbf{} & \textbf{process} & \textbf{avg (5-shot)} & \textbf{ score } & \textbf{avg} & \textbf{avg} & \textbf{date} \\
\midrule
Mistral-7B-v0.2 & pre-training & 38.81 & - & - & 60.78 & Mar 2024 \\
Bielik-11B-v2 & cont pre-training & 58.14 & - & 15.99 & 65.87 & Aug 2024 \\
Bielik-11B-v2.0-Instruct & SFT & 64.98 & 7.56 & 24.66 & 68.04 & Aug 2024 \\
Bielik-11B-v2.1-Instruct & SFT, DPOP & 65.45 & 8.00 & 27.20 & 69.82 & Aug 2024 \\
Bielik-11B-v2.2-Instruct & SFT, DPOP & 65.57 & 8.12 & 27.98 & 69.86 & Aug 2024 \\
Bielik-11B-v2.3-Instruct & SFT, DPOP, MERGE & 65.71 & 8.56 & 28.33 & 69.82 & Aug 2024 \\
Bielik-11B-v2.5-Insturct & SFT, DPOP, MERGE & 63.95 & - & - & 71.42 & May 2025 \\
\bottomrule
\end{tabular}
\caption{List of released v2 family models with source model for base line reference}
\label{tab:model-list-tab}
\end{table*}

\section*{Acknowledgements}

We gratefully acknowledge Polish high-performance computing infrastructure PLGrid (HPC Center: ACK Cyfronet AGH) for providing computer facilities and support within computational grant no. PLG/2024/017214 and PLG/2025/018338.

The model could not have been created without the commitment and work of the entire SpeakLeash team, whose contribution is invaluable. Thanks to the hard work of many individuals, it was possible to gather a large amount of content in Polish and establish collaboration between the open-science SpeakLeash project and the HPC center: ACK Cyfronet AGH. Individuals who contributed to the creation of the model through their commitment to the open-science SpeakLeash project: Sebastian Kondracki, Marek Magryś, Szymon Mazurek, Mieszko Cholewa, Igor Ciuciura, Paweł Kiszczak, Szymon Baczyński, Jacek Chwiła, Maria Filipkowska, Jan Maria Kowalski, Dominika Basaj, Kuba Sołtys, Karol Jezierski, Kacper Milan, Jan Sowa, Len Krawczyk, Marta Seidler, Agnieszka Ratajska, Krzysztof Koziarek, Szymon Pepliński, Zuzanna Dabić, Filip Bogacz, Agnieszka Kosiak, Izabela Babis, Nina Babis, and many other wonderful researchers and enthusiasts of the AI world.

\appendix
\section{European LLM Leaderboard - Additional Results}
\label{app:european-llm}

This appendix presents detailed results from the European LLM Leaderboard for German and Czech languages, complementing the Polish language results presented in the main text.

\subsection{German Language Results}

\begin{table}[h]
\centering
\small
\begin{tabular}{lrcccccc}
\toprule
\textbf{Model} & \textbf{Params} & \textbf{Avg} & \textbf{ARC} & \textbf{GSM8K} & \textbf{HellaSwag} & \textbf{MMLU} & \textbf{TruthfulQA} \\
\midrule
Meta-Llama-3.1-70B-Instruct & 70.6 & 0.71 & 0.73 & 0.74 & 0.75 & 0.78 & 0.57 \\
Gemma-2-27b-Instruct & 27.0 & 0.71 & 0.77 & 0.78 & 0.72 & 0.69 & 0.59 \\
Phi-3-medium-14B-4k-Instruct & 14.0 & 0.69 & 0.71 & 0.78 & 0.69 & 0.72 & 0.56 \\
Phi-3-medium-14B-128k-Instruct & 14.0 & 0.67 & 0.70 & 0.72 & 0.68 & 0.71 & 0.55 \\
Mixtral-8x7B-Instruct-v0.1 & 46.7 & 0.67 & 0.71 & 0.58 & 0.75 & 0.67 & 0.65 \\
Mistral-Nemo-Instruct-12.2B & 12.2 & 0.65 & 0.69 & 0.65 & 0.70 & 0.64 & 0.56 \\
c4ai-command-r-35B-v01 & 35.0 & 0.63 & 0.69 & 0.52 & 0.75 & 0.61 & 0.55 \\
Mixtral-8x7B-v0.1 & 46.7 & 0.62 & 0.70 & 0.51 & 0.74 & 0.68 & 0.50 \\
\textbf{Bielik-11B-v2.3-Instruct} & \textbf{11.2} & \textbf{0.62} & \textbf{0.64} & \textbf{0.65} & \textbf{0.62} & \textbf{0.60} & \textbf{0.59} \\
Mistral-Nemo-Base-12.2B & 12.0 & 0.61 & 0.68 & 0.52 & 0.72 & 0.65 & 0.50 \\
Gemma-2-9b-Instruct & 9.2 & 0.60 & 0.71 & 0.49 & 0.63 & 0.61 & 0.58 \\
Meta-Llama-3-8B-Instruct & 8.0 & 0.59 & 0.6 & 0.63 & 0.58 & 0.58 & 0.55 \\
Meta-Llama-3.1-8B-Instruct & 8.0 & 0.59 & 0.61 & 0.58 & 0.62 & 0.62 & 0.52 \\
EuroLLM-9B-Instruct & 9.2 & 0.58 & 0.69 & 0.47 & 0.69 & 0.58 & 0.50 \\
Mistral-7B-Instruct-v0.3 & 7.2 & 0.55 & 0.62 & 0.39 & 0.61 & 0.54 & 0.57 \\
Mistral-7B-Instruct-v0.2 & 7.2 & 0.54 & 0.61 & 0.32 & 0.61 & 0.53 & 0.64 \\
Meta-Llama-3.1-8B & 8.0 & 0.54 & 0.59 & 0.4 & 0.63 & 0.59 & 0.5 \\
Teuken-7B-sigma-v05 & 7 & 0.53 & 0.62 & 0.31 & 0.69 & 0.47 & 0.58 \\
Pharia-1-LLM-7B-control-aligned & 7.5 & 0.48 & 0.62 & 0.16 & 0.67 & 0.48 & 0.50 \\
Vicuna-13b-v1.5 & 13.0 & 0.48 & 0.55 & 0.26 & 0.59 & 0.50 & 0.52 \\
Teuken-7B-instruct-research-v0.4 & 7.5 & 0.48 & 0.62 & 0.13 & 0.64 & 0.45 & 0.54 \\
Teuken-7B-instruct-commercial-v0.4 & 7.5 & 0.46 & 0.61 & 0.09 & 0.64 & 0.45 & 0.50 \\
Salamandra-7b-instruct & 7.8 & 0.45 & 0.61 & 0.09 & 0.64 & 0.47 & 0.45 \\
Pharia-1-LLM-7B-control & 7.5 & 0.45 & 0.62 & 0.06 & 0.66 & 0.45 & 0.46 \\
Teuken-7B-base-v0.55 & 7.5 & 0.44 & 0.57 & 0.09 & 0.64 & 0.45 & 0.45 \\
Salamandra-7b & 7.8 & 0.42 & 0.60 & 0.00 & 0.64 & 0.43 & 0.45 \\
EuroLLM-1.7B-Instruct & 1.7 & 0.36 & 0.50 & 0.09 & 0.49 & 0.27 & 0.45 \\
EuroLLM-1.7B & 1.7 & 0.34 & 0.48 & 0.04 & 0.48 & 0.26 & 0.43 \\
\bottomrule
\end{tabular}
\caption{European LLM Leaderboard results for German language.}
\label{tab:european-german}
\end{table}

In German language evaluation (Table~\ref{tab:european-german}), Bielik-11B-v2.3-Instruct shows strong performance with an average score of 0.62, positioning it in the upper range among models with similar parameter counts. Despite being primarily trained on Polish and English data, the model demonstrates impressive cross-lingual transfer capabilities to German. It achieves particularly strong results in mathematical reasoning (GSM8K: 0.65) and factual accuracy (TruthfulQA: 0.59), outperforming several models with similar or larger parameter counts such as Mistral-7B-Instruct-v0.3 (0.55), Meta-Llama-3.1-8B (0.54), and even Vicuna-13b-v1.5 (0.48). This performance indicates Bielik's effectiveness in transferring its capabilities to Germanic languages despite its Slavic-focused training.

\subsection{Czech Language Results}

\begin{table}[h]
\centering
\small
\begin{tabular}{lrcccccc}
\toprule
\textbf{Model} & \textbf{Params} & \textbf{Avg} & \textbf{ARC} & \textbf{GSM8K} & \textbf{HellaSwag} & \textbf{MMLU} & \textbf{TruthfulQA} \\
\midrule
Meta-Llama-3.1-70B-Instruct & 70.6 & 0.71 & 0.72 & 0.75 & 0.72 & 0.77 & 0.58 \\
Gemma-2-27b-Instruct & 27.4 & 0.70 & 0.75 & 0.75 & 0.70 & 0.68 & 0.61 \\
Mixtral-8x7B-Instruct-v0.1 & 46.7 & 0.62 & 0.68 & 0.50 & 0.66 & 0.63 & 0.63 \\
c4ai-command-r-35B-v01 & 35.0 & 0.60 & 0.67 & 0.48 & 0.71 & 0.60 & 0.55 \\
\textbf{Bielik-11B-v2.3-Instruct} & \textbf{11.2} & \textbf{0.60} & \textbf{0.63} & \textbf{0.60} & \textbf{0.59} & \textbf{0.59} & \textbf{0.58} \\
Mistral-Nemo-Instruct-12.2B & 12.2 & 0.59 & 0.62 & 0.55 & 0.59 & 0.59 & 0.59 \\
Mixtral-8x7B-v0.1 & 46.7 & 0.58 & 0.66 & 0.43 & 0.65 & 0.65 & 0.52 \\
Gemma-2-9b-Instruct & 9.0 & 0.58 & 0.67 & 0.46 & 0.59 & 0.59 & 0.60 \\
Meta-Llama-3.1-8B-Instruct & 8.0 & 0.57 & 0.58 & 0.58 & 0.57 & 0.58 & 0.53 \\
EuroLLM-9B-Instruct & 9.0 & 0.57 & 0.68 & 0.45 & 0.67 & 0.58 & 0.50 \\
Mistral-7B-Instruct-v0.3 & 7.0 & 0.50 & 0.58 & 0.32 & 0.54 & 0.51 & 0.56 \\
Teuken-7B-sigma-v05 & 7.0 & 0.50 & 0.59 & 0.26 & 0.65 & 0.45 & 0.56 \\
Mistral-7B-Instruct-v0.2 & 7.0 & 0.50 & 0.57 & 0.28 & 0.55 & 0.50 & 0.59 \\
Teuken-7B-instruct-research-v0.4 & 7.0 & 0.45 & 0.59 & 0.11 & 0.62 & 0.42 & 0.53 \\
Salamandra-7b-instruct & 7.0 & 0.45 & 0.59 & 0.11 & 0.62 & 0.46 & 0.47 \\
Teuken-7B-instruct-commercial-v0.4 & 7.0 & 0.44 & 0.56 & 0.11 & 0.61 & 0.43 & 0.51 \\
Vicuna-13b-v1.5 & 13.0 & 0.43 & 0.48 & 0.20 & 0.51 & 0.46 & 0.49 \\
Salamandra-7b & 7.0 & 0.41 & 0.59 & 0.00 & 0.63 & 0.41 & 0.43 \\
Occiglot-7b-eu5 & 7.0 & 0.40 & 0.46 & 0.18 & 0.48 & 0.44 & 0.45 \\
Occiglot-7b-eu5-Instruct & 7.0 & 0.40 & 0.48 & 0.13 & 0.48 & 0.43 & 0.46 \\
EuroLLM-1.7B-Instruct & 2.0 & 0.35 & 0.49 & 0.08 & 0.48 & 0.26 & 0.44 \\
EuroLLM-1.7B & 2.0 & 0.33 & 0.46 & 0.03 & 0.48 & 0.25 & 0.43 \\
Pharia-1-LLM-7B-control & 7.0 & 0.27 & 0.27 & 0.01 & 0.31 & 0.30 & 0.44 \\
Pharia-1-LLM-7B-control-aligned & 7.0 & 0.27 & 0.27 & 0.02 & 0.31 & 0.31 & 0.43 \\
\bottomrule
\end{tabular}
\caption{European LLM Leaderboard results for Czech language.}
\label{tab:european-czech}
\end{table}

In Czech language evaluation (Table~\ref{tab:european-czech}), Bielik-11B-v2.3-Instruct achieves a strong average score of 0.60, placing it fifth overall and outperforming all models with comparable parameter counts. This demonstrates exceptional cross-lingual transfer to Czech, another West Slavic language. The model's performance is particularly impressive in mathematical problem-solving with a GSM8K score of 0.60, surpassing much larger models including Mixtral-8x7B-Instruct-v0.1 (0.50) and c4ai-command-r-35B-v01 (0.48). 

Bielik significantly outperforms all 7B and smaller models, including Mistral-7B-Instruct-v0.3 (0.50), specialized European models like Teuken and Salamandra, and even Vicuna-13b-v1.5 (0.43) despite its larger parameter count. The factual accuracy score (TruthfulQA: 0.58) matches that of Meta-Llama-3.1-70B-Instruct despite Bielik having only about 15\% of its parameters. These results highlight Bielik's efficient architecture and the benefits of its Polish pretraining for cross-lingual transfer to related Slavic languages.

\subsection{FLORES200 Translation Benchmark Results}

\begin{table}[h]
\centering
\small
\begin{tabular}{lrr}
\toprule
\textbf{Language} & \textbf{to Polish} & \textbf{from Polish} \\
\midrule
Bulgarian & 17.83 & 6.81 \\
Czech & 19.30 & 14.58 \\
Danish & 17.81 & 15.24 \\
German & 19.18 & 14.93 \\
Greek & 5.49 & 0.50 \\
English & 21.93 & 28.32 \\
Estonian & 6.13 & 1.53 \\
Finnish & 12.76 & 4.14 \\
French & 18.97 & 19.06 \\
Hungarian & 15.59 & 7.36 \\
Italian & 16.54 & 13.07 \\
Lithuanian & 7.99 & 1.28 \\
Latvian & 5.85 & 0.88 \\
Dutch & 14.76 & 13.07 \\
Portuguese & 19.10 & 19.76 \\
Romanian & 18.30 & 15.47 \\
Slovak & 17.65 & 6.60 \\
Slovenian & 17.37 & 12.53 \\
Spanish & 16.17 & 17.50 \\
Swedish & 17.58 & 14.51 \\
\midrule
Average & 15.31 & 11.36 \\
\bottomrule
\end{tabular}
\caption{FLORES200 translation benchmark BLEU scores for Bielik-11B-v2.3-Instruct}
\label{tab:flores-translation}
\end{table}

The FLORES200 translation results for Bielik-11B-v2.3-Instruct (Table~\ref{tab:flores-translation}) show asymmetric performance across language pairs with Polish. The model achieves its best results with English-Polish (21.93/28.32) and demonstrates strong performance with linguistically similar West Slavic languages like Czech (19.30/14.58). It also performs well with major European languages such as Portuguese (19.10/19.76), French (18.97/19.06), and German (19.18/14.93). However, performance drops significantly with Baltic and Finno-Ugric languages, suggesting limitations in cross-linguistic transfer to more distant language families.

\section{BenCzechMark Benchmark Results}
\label{app:benczechmark}

BenCzechMark (BCM) is a comprehensive Czech-centric benchmark for Large Language Models introduced by \cite{fajcik2024benczechmarkczechcentricmultitask}. It encompasses 50 challenging tasks across 8 categories, evaluating models on diverse linguistic capabilities in native Czech. The benchmark's scoring system employs a Duel Win Score (DWS) mechanism that statistically compares each model against others, providing a robust evaluation methodology.

\begin{table}[h]
\centering
\small
\begin{tabular}{lrrrrrrrrrr}
\toprule
\textbf{Model} & \textbf{Params} & \textbf{Avg} & \textbf{Czech} & \textbf{Math} & \textbf{Factual} & \textbf{Lang.} & \textbf{NER} & \textbf{NLI} & \textbf{Reading} & \textbf{Sentiment} \\
 & \textbf{(B)} & \textbf{(\%)} & \textbf{Lang.} & \textbf{Reason.} & \textbf{Know.} & \textbf{Model.} &  &  & \textbf{Comp.} &  \\
\midrule
Meta-Llama-3.1-405B-Instruct & 406.0 & 90.2 & 98.5 & 91.7 & 88.6 & 93.3 & 97.2 & 89.1 & 93.2 & 70.4 \\
Qwen2.5-72B & 72.7 & 85.7 & 75.3 & 94.9 & 83.6 & 83.6 & 96.3 & 84.0 & 92.6 & 75.5 \\
Meta-Llama-3.3-70B-Instruct & 70.6 & 80.0 & 86.1 & 75.9 & 78.8 & 69.9 & 92.6 & 82.8 & 79.0 & 75.0 \\
Qwen2.5-72B-Instruct & 72.7 & 78.9 & 80.2 & 94.0 & 80.7 & 69.0 & 69.4 & 86.1 & 72.8 & 79.2 \\
Meta-Llama-3.1-70B-Instruct & 70.6 & 78.6 & 80.2 & 60.2 & 79.1 & 81.7 & 88.9 & 84.6 & 78.4 & 75.9 \\
Meta-Llama-3.1-70B & 70.6 & 76.5 & 70.1 & 55.1 & 75.9 & 90.5 & 97.2 & 68.8 & 83.3 & 71.3 \\
Qwen2.5-32B-Instruct & 32.8 & 73.9 & 74.7 & 91.7 & 68.0 & 62.0 & 65.7 & 84.0 & 68.5 & 76.9 \\
Mistral-Large-Instruct-2411 & 123.0 & 73.8 & 85.2 & 83.3 & 86.5 & 6.5 & 82.4 & 83.4 & 82.1 & 81.0 \\
Qwen2-72B-Instruct & 72.7 & 72.5 & 72.2 & 83.3 & 78.6 & 60.6 & 80.6 & 66.1 & 74.1 & 64.4 \\
Mixtral-8x22B-Instruct-v0.1 & 141.0 & 71.5 & 69.1 & 59.3 & 70.6 & 83.3 & 70.4 & 73.5 & 67.3 & 78.2 \\
\textbf{Bielik-11B-v2.3-Instruct} & \textbf{11.2} & \textbf{49.5} & \textbf{46.9} & \textbf{31.0} & \textbf{48.7} & \textbf{25.9} & \textbf{54.6} & \textbf{67.1} & \textbf{64.2} & \textbf{57.4} \\
Google/Gemma-2-27b & 27.2 & 46.9 & 51.2 & 52.3 & 43.7 & 63.4 & 2.8 & 53.0 & 38.9 & 69.9 \\
Meta-Llama-3-8B-Instruct & 8.0 & 44.0 & 45.4 & 13.0 & 35.2 & 33.6 & 63.0 & 48.4 & 56.2 & 57.4 \\
EuroLLM-9B-Instruct & 9.2 & 34.7 & 43.8 & 11.6 & 36.2 & 85.2 & 14.8 & 29.3 & 29.6 & 27.3 \\
EuroLLM-9B & 9.2 & 31.4 & 32.4 & 8.8 & 36.0 & 88.7 & 16.7 & 24.5 & 23.5 & 20.4 \\
Mistral-7B-Instruct-v0.3 & 7.3 & 32.6 & 29.0 & 4.2 & 33.1 & 37.3 & 38.9 & 39.9 & 35.8 & 42.6 \\
CohereForAI/aya-23-8B & 8.0 & 37.4 & 27.2 & 5.6 & 31.0 & 53.5 & 40.7 & 34.0 & 45.1 & 62.0 \\
salamandra-7b-instruct & 7.8 & 29.4 & 24.1 & 2.8 & 25.1 & 70.1 & 11.1 & 29.5 & 16.7 & 56.0 \\
csmpt7b & 6.7 & 16.0 & 4.6 & 0.0 & 2.6 & 80.6 & 9.3 & 18.5 & 3.1 & 9.3 \\
CSTinyLlama-1.2B & 1.2 & 11.3 & 1.2 & 1.4 & 2.9 & 76.2 & 4.6 & 2.8 & 0.0 & 1.4 \\
\bottomrule
\end{tabular}
\caption{BenCzechMark results showing performance across all 8 categories. The table includes top-performing models, various mid-sized models (including Bielik-11B-v2.3-Instruct ranked 21st overall), and specialized Czech models for comparison. All values are percentages, with higher scores indicating better performance.}
\label{tab:benczechmark-comprehensive}
\end{table}

Bielik-11B-v2.3-Instruct achieves an average score of 49.5\% on the BenCzechMark leaderboard (Table~\ref{tab:benczechmark-comprehensive}), placing it at position 21 among evaluated models. This is a respectable performance for a model of its size (11.2B parameters) when compared to much larger models like Llama-3.1-405B-Instruct (406B) and Qwen2.5-72B (72.7B) that dominate the top positions.

\paragraph{Category-specific performance:} Bielik shows varying capabilities across different task categories:
\begin{itemize}
    \item Strongest performance in Natural Language Inference (67.1\%) and Reading Comprehension (64.2\%), demonstrating good reasoning abilities in Czech
    \item Moderate performance in Sentiment Analysis (57.4\%) and NER (54.6\%), showing capability in understanding emotional context and entity recognition in Czech text
    \item Lower scores in Czech Language Understanding (46.9\%), Factual Knowledge (48.7\%), and Math Reasoning (31.0\%), indicating areas for improvement in future versions
\end{itemize}

\paragraph{Comparative analysis:} Despite being primarily designed for Polish language processing, Bielik demonstrates reasonable cross-lingual transfer to Czech:
\begin{itemize}
    \item Outperforms several similarly-sized models in Reading Comprehension and NLI tasks
    \item Shows stronger reasoning capabilities in NLI compared to some larger models
    \item Significantly outperforms native Czech models (BUT-FIT/csmpt7b and BUT-FIT/CSTinyLlama-1.2B) in overall performance and most categories, despite these models having been specifically trained on Czech data

    \item Demonstrates the value of instruction tuning and Slavic language pretraining for cross-lingual transfer between related languages
\end{itemize}

These results highlight Bielik's ability to generalize across Slavic languages, while also identifying specific areas where future versions could be enhanced to better support Czech language capabilities. The model's performance is particularly notable given that it was not specifically optimized for Czech language tasks, unlike the specialized Czech models in the benchmark that excel in language modeling but lag significantly in reasoning-based tasks.

\section{Portuguese Benchmark Results}
\label{app:portuguese}

\begin{table}[h]
\centering
\small
\begin{tabular}{lcccccccccc}
\toprule
\textbf{Model} & \textbf{Avg} & \textbf{ENEM} & \textbf{BLUEX} & \textbf{OAB} & \textbf{ASSIN2} & \textbf{ASSIN2} & \textbf{FAQUAD} & \textbf{HateBR} & \textbf{PT Hate} & \textbf{tweetSent} \\
 &  &  &  & \textbf{Exams} & \textbf{RTE} & \textbf{STS} & \textbf{NLI} &  & \textbf{Speech} & \textbf{BR} \\
\midrule
Bielik-11B-v2.1-Instruct & 73.45 & 72.29 & 62.59 & 48.56 & 93.50 & 79.82 & 82.73 & 79.58 & 73.97 & 67.99 \\
Bielik-11B-v2.2-Instruct & 73.45 & 72.57 & 62.73 & 49.02 & 93.67 & 80.50 & 80.32 & 79.60 & 73.75 & 68.85 \\
Bielik-11B-v2.3-Instruct & 73.08 & 71.73 & 62.31 & 48.66 & 93.29 & 80.49 & 80.63 & 79.58 & 73.13 & 67.89 \\
Bielik-11B-v2.0-Instruct & 72.57 & 70.75 & 61.47 & 47.84 & 92.59 & 80.14 & 83.75 & 77.88 & 72.06 & 66.61 \\
Bielik-11B-v2 & 68.68 & 68.86 & 58.55 & 48.70 & 92.76 & 77.99 & 61.37 & 74.05 & 67.26 & 68.58 \\
Mistral-Nemo-Instruct-2407 & 73.76 & 72.08 & 61.47 & 53.80 & 93.22 & 75.18 & 82.34 & 77.96 & 74.30 & 73.48 \\
\bottomrule
\end{tabular}
\caption{Performance of Bielik models on Portuguese language benchmarks (higher is better). The table shows scores for: ENEM (Brazilian National High School Exam), BLUEX, OAB Exams (Brazilian Bar Association), ASSIN2 RTE (Recognizing Textual Entailment), ASSIN2 STS (Semantic Textual Similarity), FAQUAD NLI (Natural Language Inference), HateBR, PT Hate Speech, and tweetSentBR.}
\label{tab:portuguese-benchmark}
\end{table}

The Open PT LLM Leaderboard \cite{open-pt-llm-leaderboard} aims to provide a comprehensive benchmark for the evaluation of Large Language Models (LLMs) in the Portuguese language across a variety of tasks and datasets. The benchmark evaluates models on 9 key tasks using the Eleuther AI Language Model Evaluation Harness, a unified framework to test generative language models on a large number of different evaluation tasks. As shown in Table \ref{tab:portuguese-benchmark}, Bielik models perform competitively on Portuguese language tasks despite being primarily optimized for Polish, demonstrating strong cross-lingual transfer abilities to other Romance languages.

\bibliographystyle{unsrtnat}
\bibliography{references} 

\end{document}